%% file: acl_latex.tex
\pdfoutput=1
\documentclass[11pt]{article}

\usepackage[preprint]{acl}

\usepackage{times}
\usepackage{latexsym}

\usepackage[T1]{fontenc}
\usepackage[utf8]{inputenc}

\usepackage{microtype}

\usepackage{inconsolata}

\usepackage[utf8]{inputenc} % allow utf-8 input
\usepackage[T1]{fontenc}    % use 8-bit T1 fonts
\usepackage{hyperref}       % hyperlinks
\usepackage{url}            % simple URL typesetting
\usepackage{booktabs}       % professional-quality tables
\usepackage{amsfonts}       % blackboard math symbols
\usepackage{nicefrac}       % compact symbols for 1/2, etc.
\usepackage{microtype}      % microtypography
\usepackage{xcolor}         % colors
\usepackage{multirow}
\usepackage{amsmath}
\usepackage{bbm}
\usepackage{graphicx}
\usepackage{wrapfig}
\usepackage{enumitem}
\usepackage{longtable}
\usepackage{colortbl}
\usepackage{longtable}
\usepackage{pifont}
\usepackage{inconsolata}
\DeclareMathOperator*{\argmin}{argmin}

\title{LEMoE: Advanced Mixture of Experts Adaptor for Lifelong Model Editing of Large Language Models}

\author{Renzhi Wang, Piji Li$^{\ast}$\\ % All authors must be in the same font size and format. Use \Large and \textbf to achieve this result when breaking a line
College of Computer Science and Technology,\\
Nanjing University of Aeronautics and Astronautics, China\\
MIIT Key Laboratory of Pattern Analysis and Machine Intelligence, Nanjing, China\\
\texttt{\{rzhwang,pjli\}@nuaa.edu.cn}}

\begin{document}
\maketitle

\renewcommand{\thefootnote}{\fnsymbol{footnote}}
\footnotetext[1]{Corresponding author}
\renewcommand{\thefootnote}{\arabic{footnote}}

\begin{abstract}
Large language models (LLMs) require continual knowledge updates to stay abreast of the ever-changing world facts, prompting the formulation of lifelong model editing task. While recent years have witnessed the development of various techniques for single and batch editing, these methods either fail to apply or perform sub-optimally when faced with lifelong editing. 
In this paper, we introduce LEMoE, an advanced Mixture of Experts (MoE) adaptor for lifelong model editing. We first analyze the factors influencing the effectiveness of conventional MoE adaptor in lifelong editing, including catastrophic forgetting, inconsistent routing and order sensitivity. Based on these insights, we propose a tailored module insertion method to achieve lifelong editing, incorporating a novel KV anchor routing to enhance routing consistency between training and inference stage, along with a concise yet effective clustering-based editing order planning. 
Experimental results demonstrate the effectiveness of our method in lifelong editing, surpassing previous model editing techniques while maintaining outstanding performance in batch editing task. Our code will be available.
\end{abstract}

\section{Introduction}
Large language models~\cite{OpenAI, llama, LLama2, mixtral, qwen} encode a vast amount of world knowledge during pre-training, which can be accessed and utilized through natural language prompts~\cite{DBLP:conf/emnlp/PetroniRRLBWM19}. However, the dynamic nature of the real world necessitates regular and continual updates to these models to correct outdated information or integrate new knowledge~\cite{SCEN, wise}. Also, retraining or fine-tuning of LLMs is often resource-intensive and time-consuming~\cite{hook_layer}, making it impractical for lifelong growing knowledge. Therefore, lifelong model editing~\cite{GRACE} has been proposed to remedy the continual knowledge updates and injections for LLMs in a cheap and timely manner~\cite{wise}.

\begin{figure}
\centering
\includegraphics[width=\linewidth]{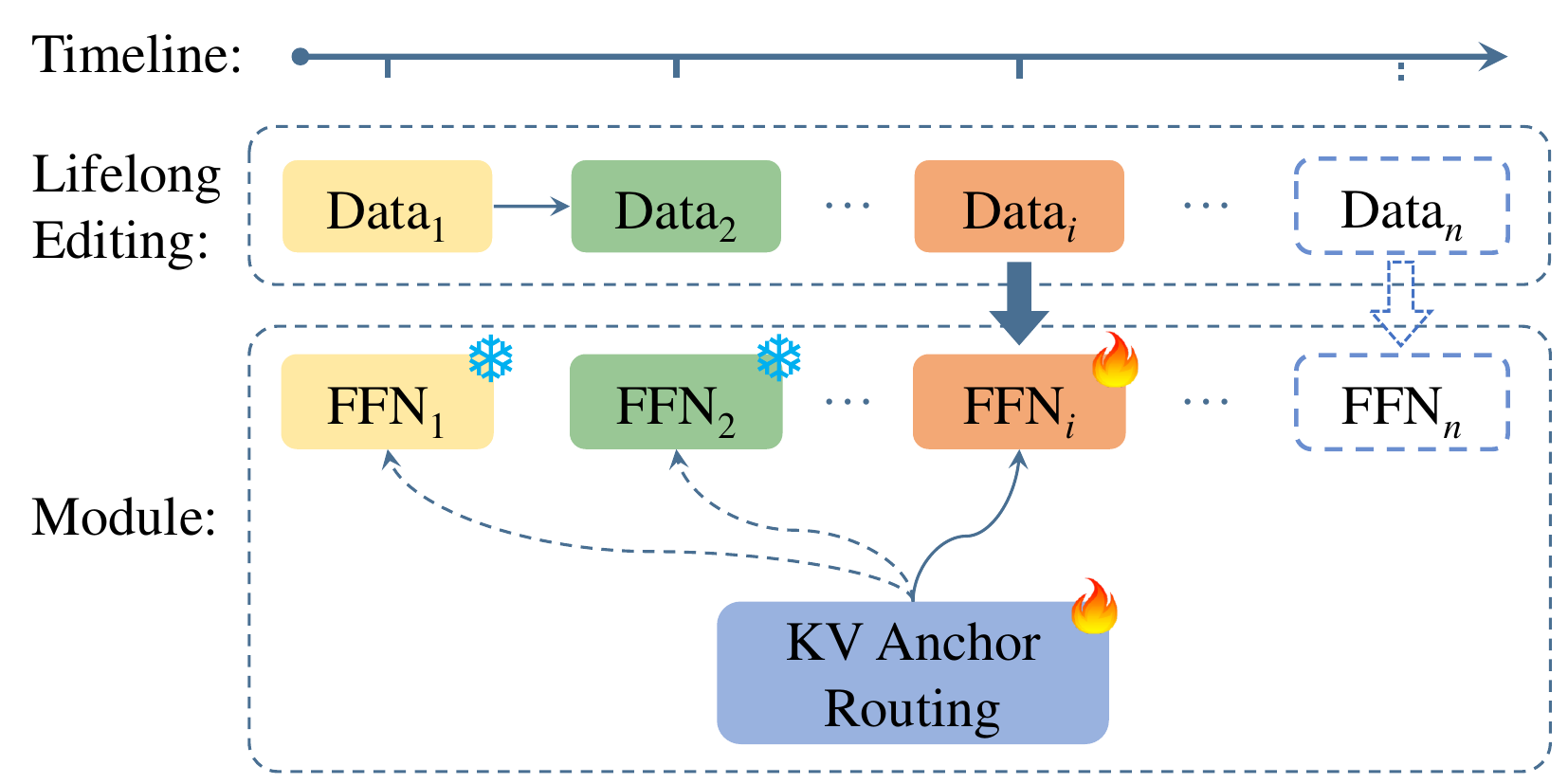}
\caption{The conceptual framework for LEMoE. We align the expert networks in MoE architecture with data batches and freeze the expert networks corresponding to previous data when conducting current edits. $\text{Data}_i$ and $\text{FFN}_i$ represent the current data and module, with dashed line parts indicating future edits.} 
\label{fig:intro_fig}
%\vspace{-3mm}
\end{figure}

In recent years, there has been a proliferation of effective model editing techniques proposed for single or batch editing, such as MEND~\cite{mend}, ROME~\cite{rome}, MEMIT~\cite{MEMIT}, and MEMoE~\cite{memoe}. However, these methods often prove inapplicable or exhibit suboptimal performance when faced with lifelong editing task~\cite{wise}. In this paper, we introduce LEMoE, an advanced Mixture of Experts (MoE) adaptor, to address the challenges inherent in lifelong editing. 

Initially, we analyze the factors that influence the effectiveness of conventional MoE adaptor in lifelong editing, including catastrophic forgetting, inconsistent routing, and order sensitivity. 
In the Catastrophic Forgetting Analysis (\S \ref{subsec:catastrophic}), we evaluate the performance of conventional MoE adaptor at different positions within the editing sequence to quantify the impact of subsequent edits on preceding ones. We observe that the conventional MoE adaptor exhibits significant catastrophic forgetting, where earlier edits are more prone to errors. 
In the Routing Consistency Analysis (\S \ref{subsec:routing consistency}), we compare the expert networks selected by routing strategy during the training and inference stages when faced with the same input. This comparison reveals a routing inconsistency in the conventional routing strategy, where identical inputs are routed to different experts at different stages. 
Finally, in the Order Sensitivity Analysis, we highlight that editing order profoundly impacts model performance (\S \ref{subsec:Order Sensitivity Analysis}). 
Through varying the sequence order of the same dataset during lifelong editing, we observe performance variations of up to 20 points, surpassing the improvement of some optimization algorithms.

Based on these insights, we propose a tailored module insertion method to achieve lifelong editing (\S \ref{subsec:New Module Inserting}). As illustrated in Figure \ref{fig:intro_fig}, we align the expert networks in the MoE architecture with the data batches in the sequential editing process. When conducting current editing, we freeze the expert network corresponding to the previous data, thereby mitigating the adverse effects of current data editing on previous edits and alleviating catastrophic forgetting from a model mechanism perspective.
Secondly, we introduce a novel Key-Value (KV) anchor routing (\S \ref{subsec:KV Anchor Routing}), wherein each expert is assigned a key vector and the input instance-level embedding serves as the corresponding value. Based on these key-value pairs, we align the routing computation processes during both training and inference stage. This ensures that identical inputs undergo same routing computation to reach the same expert across all stages, thereby enhancing routing consistency and further mitigating catastrophic forgetting.
Finally, leveraging the consistency between the MoE preferences of editing order and the objectives of clustering algorithm, we employ a concise yet effective clustering-based order planning to enhance the overall performance of LEMoE (\S \ref{subsec:Clustering-based Order Planning}).

We conduct experiments on the LLaMA-7B and Mistral-7B models using the ZsRE~\cite{ZsRE} and SelfCheckGPT~\cite{selfcheckgpt} datasets to evaluate the performance of LEMoE. Experimental results show that our approach surpasses previous model editing methods, while maintaining excellent performance in batch editing. 

The main contributions of our work can be summarized as follows:
\vspace{-2mm}
\begin{itemize}[leftmargin=0.5cm]
        \setlength\itemsep{0em}
    \item We analyze the influential factors of conventional MoE adaptor in lifelong editing task, including catastrophic forgetting, inconsistent routing, and order sensitivity.
    \item We introduce LEMoE, an advanced MoE adaptor for lifelong model editing. To address the aforementioned challenges, we propose a module insertion method, KV anchor routing, and clustering-based order planning.
    \item Experimental results show the efficacy of our proposed method in lifelong editing, while simultaneously preserving outstanding performance in batch editing.
\end{itemize}

\begin{figure*}
\centering
\includegraphics[width=\linewidth]{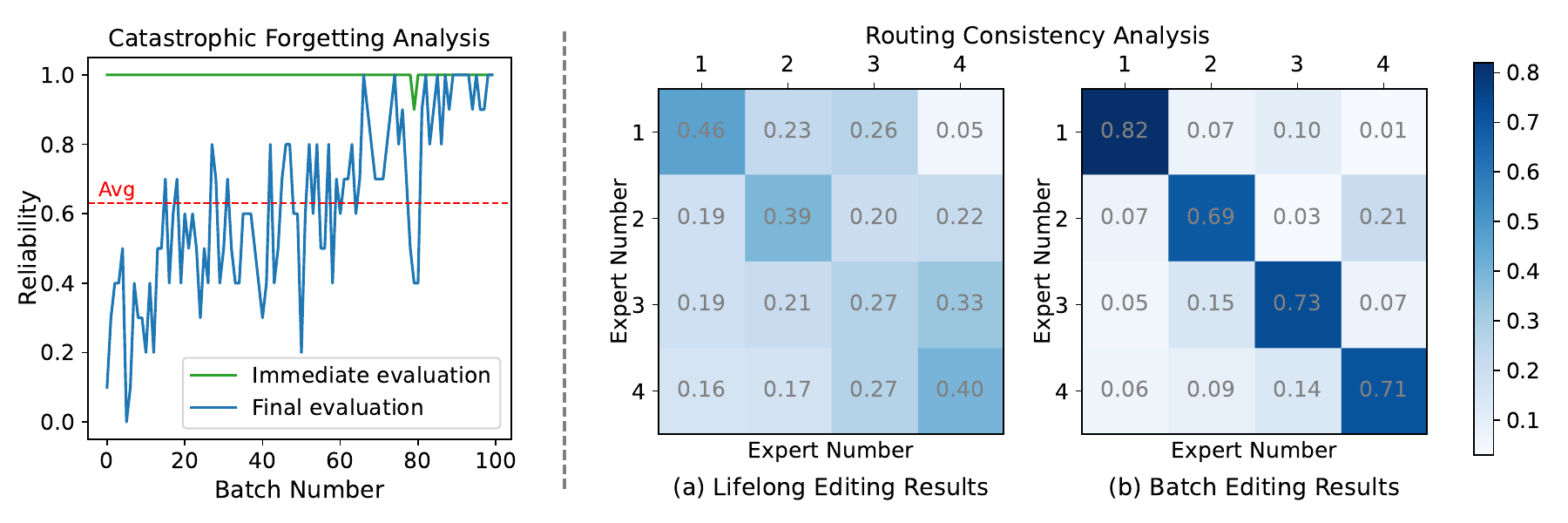}
%\vspace{-7mm}
\caption{\emph{Left:} Reliability of conventional MoE under different stage evaluation. ``Immediate evaluation'' occurs immediately after each edit, ``Final evaluation'' occurs after all edits in lifelong editing. \emph{Right:} Visualization of routing consistency. The value $C_{ij}$ in each block denotes the proportion of the input data processed by expert $i$ during the training phase that is routed to expert $j$ during the testing phase. \texttt{Model: LLaMA2-7B. Dataset: ZsRE.}}
\label{pic:factors12}
%\vspace{-3mm}
\end{figure*}

%\vspace{-2mm}
\section{Preliminaries of Model Editing}
\label{sec:preliminaries}
Based on previous research~\cite{KE_oppotunity,KE_survey, hook_layer}, model editing involves the process of transforming an initial base model $f_{\theta}$ (where $\theta$ denotes the model's parameters) into an edited model $f_{\theta^{'}}$. The goal is to modify the model's outputs for a specific set of editing instances, while maintaining consistent behavior for all other instances \cite{hook_layer}. 
The target editing instance can be described as $(x_i^e, y_i^e)$, with the condition that $f_{\theta}(x_i^e)\neq y_i^e$. The set of this instances is termed the editing scope $I_{edit}$, whereas the out-of-scope set $O_{edit}$ comprises inputs not associated with the editing examples. Formally, the criteria for a successful edit can be described as:
%\vspace{-2mm}
\begin{equation}
f_{\theta^{'}}(x_i) = \begin{cases}
y_i^e & \text{if } x_i \in I_{edit} \\
f_{\theta}(x_i) & \text{if } x_i \in O_{edit}
\end{cases}
\end{equation}
%\vspace{-4mm}

We divide model editing tasks into two categories: Batch Editing and Lifelong Editing:

%\vspace{2mm}
\noindent1) \textbf{Batch Editing} refers to the simultaneous modification of model $f_{\theta}$ using multiple input instances in one editing operation:
%\vspace{-2mm}
\begin{equation}
    \theta' \leftarrow \argmin_\theta \sum\nolimits_{i=1}^{n}(\parallel{f_{\theta}(x_i^e)-y_i^e}\parallel)\ 
\end{equation}
where $n$ represents the batch size. Batch editing with batch size of 1 is also known as Single Editing.

%\vspace{2mm}
\noindent2) \textbf{Lifelong Editing} refers to the continuous iterative modification of model $f_{\theta}$, also known as Sequential Batch Editing. Lifelong editing use dataset $\mathcal{D}_{edit}=\{\mathcal{B}_1,\mathcal{B}_2,\dots,\mathcal{B}_s\}$ with $s$ sequential batches and each batch $\mathcal{B}_i$ contains $n$ edits:
%\vspace{-3mm}
\begin{equation}
    \theta' \leftarrow \argmin_\theta \sum_{j=1}^{s} \sum_{i=(j-1) \times n + 1}^{j \times n}(\parallel{f_{\theta}(x_i^e)-y_i^e}\parallel)\ 
\end{equation}
Note that the model is not rolled back to the initial state after each batch editing. Similarly, lifelong editing with batch size of 1 in the sequence is also referred to as Sequential Editing.

Based on the above settings, an effective model editor must satisfy the criteria of three fundamental properties: Reliability, Generality, and Locality \cite{KE_oppotunity}. These properties are formally defined as follows \cite{KE_survey}:

%\vspace{2mm}
\noindent1) \textbf{Reliability} denotes the average precision of the post-edit model $f_{\theta^{'}}$ concerning the intended edits:
\begin{equation}
\mathbb{E}_{(x_i^e, y_i^e) \sim I_{edit}} \mathbbm{1} \left\{\operatorname{argmax}_y f_{\theta^{'}}\left(y \mid x_{i}^{e}\right)=y_{i}^{e}\right\}
\end{equation}

\noindent2) \textbf{Generality} quantifies the average precision of the model $f_{\theta^{'}}$ on instances uniformly sampled from the equivalence neighborhood $N_{edit}$, encompassing input/output pairs pertinent to $I_{edit}$:
\begin{equation}
\mathbb{E}_{(x_i, y_i^e) \sim N_{edit}} \mathbbm {1} \left\{\operatorname{argmax}_yf_{\theta^{'}}\left(y \mid x_{i}\right)=y_{i}^e\right\}
\end{equation} 

\noindent3) \textbf{Locality} is measured by the proportion at which predictions of the post-edit model $f_{\theta^{'}}$ remain unaltered compared to the pre-edit model $f_{\theta}$:
\begin{equation}
\mathbb{E}_{(x_i, y_i) \sim O_{edit}} \mathbbm {1} \left\{f_{\theta^{'}}\left(y \mid x_i\right)=f_{\theta}\left(y \mid x_i\right) \right\}
\end{equation}

\section{Analysis of Influencing Factors}
In this section, we analyze the factors that influence the effectiveness of conventional MoE adaptor in lifelong editing, including Catastrophic Forgetting Analysis (\S \ref{subsec:catastrophic}), Routing Consistency Analysis (\S \ref{subsec:routing consistency}), and Order Sensitivity Analysis (\S \ref{subsec:Order Sensitivity Analysis}).

\subsection{Catastrophic Forgetting Analysis}
\label{subsec:catastrophic}
In the field of continual learning, the general phenomenon of catastrophic forgetting where training on new tasks degrade performance on old tasks has been extensively reported and studied~\cite{DBLP:journals/corr/abs-2309-10105}. We aim to investigate whether the MoE adaptor in lifelong model editing also suffers from catastrophic forgetting: whether editing with new data leads to forgetting previously edited data. To assess this, we employ the classic evaluation method for catastrophic forgetting, which involves measuring the performance decrease on previously edited data during the course of lifelong editing.
%\vspace{-1mm}
\paragraph{Experiments}
To evaluate the impact of current data editing on previous ones during the lifelong editing process, we employ two different stage evaluation methods: (1) a normal evaluation conducted on all editing data only after all edits are completed, and (2) an evaluation conducted immediately after editing the current data to assess the effectiveness of these edits at the current stage. We do not set up another control group where the base model edits only the current data without considering previous data because the accuracy of MoE adaptor under the second evaluation method is nearly 100\%. 
We utilize the LLaMA2-7B as base model and ZsRE dataset (detailed in \S \ref{par:datasts and metrics}). In lifelong editing setting, we perform 100 sequential editing steps, with each step editing a batch of 10 instances, resulting in a total of 1000 edited instances. The evaluation metric is Reliability (detailed in \S \ref{sec:preliminaries}). The implementation of conventional MoE adaptor follows~\cite{memoe}, employing 4 experts and $top_k=1$.
%\vspace{-1mm}
\paragraph{Results}
In the left plot of Figure \ref{pic:factors12}, the reliability of the immediate evaluation during the entire lifelong editing process shows that only once do the score fall below 100. This indicates that the model consistently achieves desired editing goals at every individual step. However, the model's overall performance is only around 60 in the final evaluation, with earlier edits exhibiting a more significant decline in effectiveness and some initial edits scoring close to 0. This suggests a pronounced catastrophic forgetting phenomenon, where the model's forgetfulness of previous editing data markedly diminishes its overall performance in lifelong editing.

%\vspace{-1mm}
\subsection{Routing Consistency Analysis}
\label{subsec:routing consistency}
In MoE structure, the specificity of experts directly impacts model performance~\cite{DBLP:journals/jmlr/FedusZS22}. The design philosophy of MoE adaptor encourages "professional people do professional things", ensuring that the same inputs are routed to the same expert for processing during both training and testing phases~\cite{memoe}. We aim to assess the consistency of the routing within the MoE adaptor under lifelong editing setting and explore the degree of specificity among these experts.
%\vspace{-1mm}
\paragraph{Experiments}
To assess the routing consistency, we log the processing expert for each input during the training phase and compare it against the expert processing the same input during testing. We train models under batch editing setting and lifelong editing setting on identical dataset to compare routing consistency across different tasks. In the lifelong editing setup, sequential editing steps is set to 100 steps, each step editing a batch of 10 instances. Maintaining the same edited data, batch editing utilize a batch size of 1000. We employ LLaMA2-7B and ZsRE, with all other experimental settings consistent with \S \ref{subsec:catastrophic}.
%\vspace{-1mm}
\paragraph{Results}
On the right side of Figure \ref{pic:factors12}, each subgraph depicts the proportion $C_{ij}$ where the input processed by expert $i$ during the training phase is routed to expert $j$ during the testing phase. The diagonal element $C_{ii}$ represents the probability that the same input is routed to the same expert during both training and testing phases.
Experimental results comparing two editing setups reveal that routing consistency is notably poorer in lifelong editing task, with minimal specificity observed among different experts. In contrast, batch editing exhibits significant routing consistency. Hence, there is a critical need to devise more accurate and effective routing algorithm to guide expert specialization.

%\vspace{-1mm}
\subsection{Order Sensitivity Analysis}
\label{subsec:Order Sensitivity Analysis}
In continual learning, the performance of a model significantly varies based on the order of the task arrival sequence~\cite{DBLP:journals/corr/abs-2205-13323, DBLP:conf/iclr/YoonKYH20}. Previous researches on lifelong editing ignored the impact of editing order on performance. Therefore, we aim to investigate how different editing order affect the overall performance in lifelong editing. Additionally, model tend to learn similar tasks more effectively in continual learning~\cite{DBLP:journals/corr/abs-2205-13323}, and sentences with high semantic similarity often contain related knowledge. Therefore, we also aim to explore the relationship between the semantic similarity of editing inputs and the editing results.

\begin{figure}
\centering
%\vspace{-2mm}
\includegraphics[width=\linewidth]{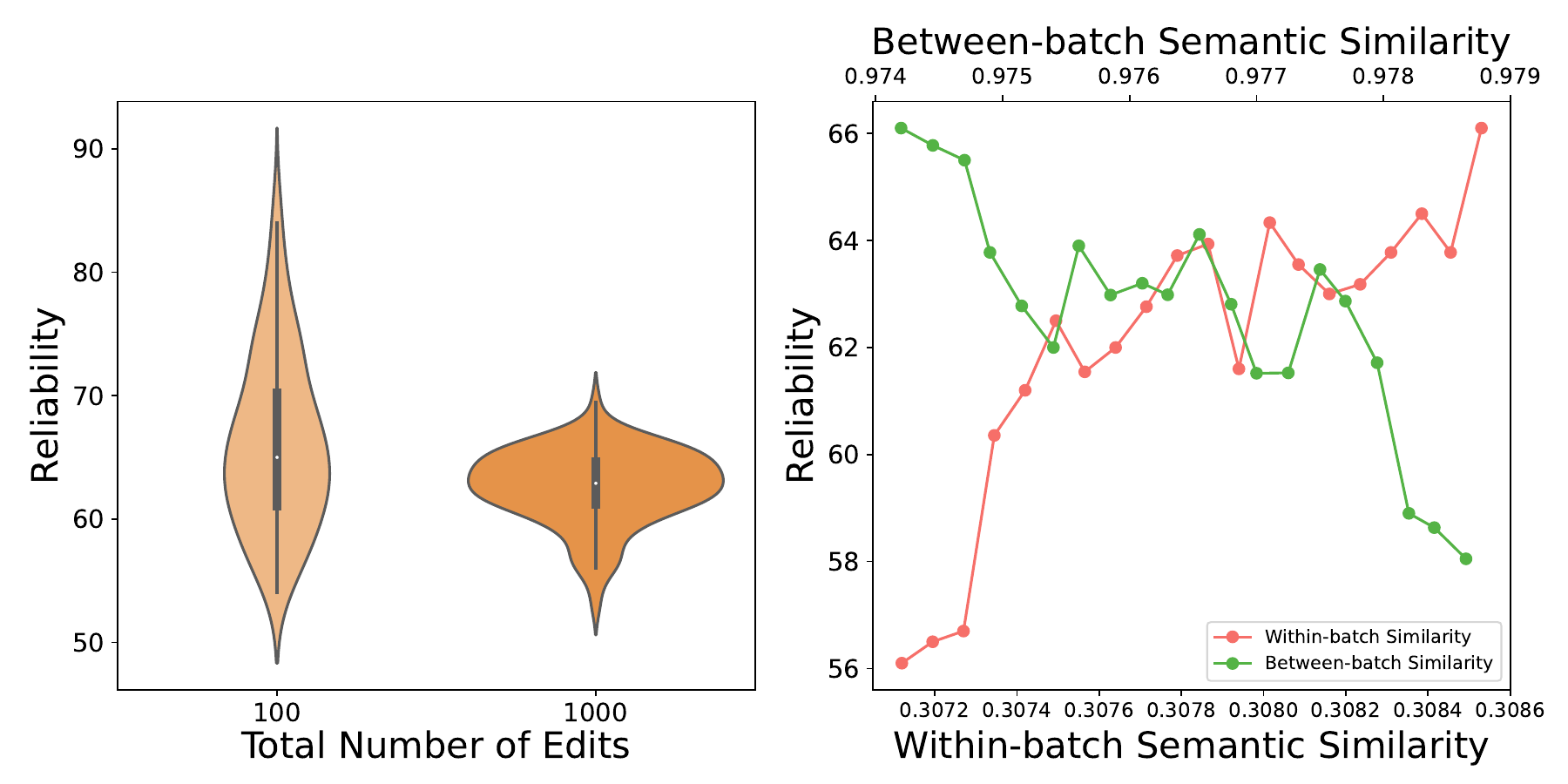}
\caption{\emph{Left:} Performance variability under different editing order. \emph{Right:} Within-Batch/Between-Batch Semantic Similarity v.s. Reliability.}
\label{pic:violine}
%\vspace{-2mm}
\end{figure}

\begin{figure*}
\centering
\includegraphics[width=\linewidth,,height=0.38\linewidth]{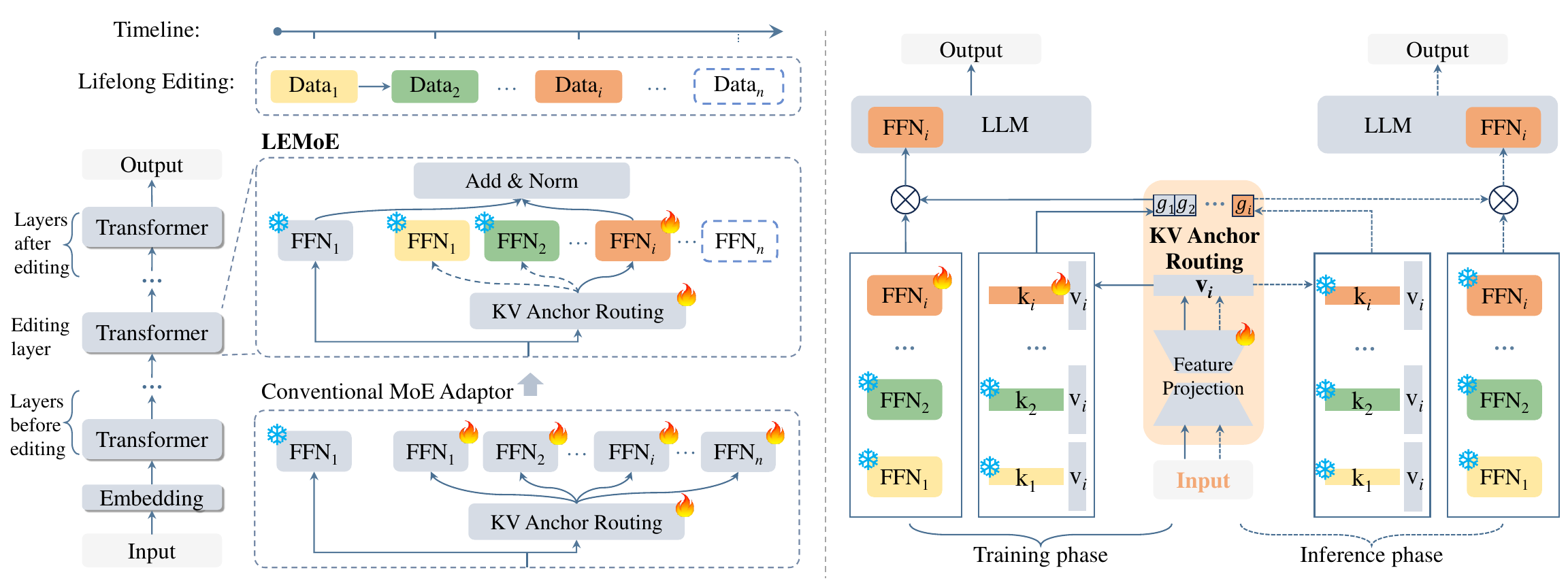}
%\vspace{-7mm}
\caption{The overall architecture of LEMoE compared with conventional MoE adaptor. We assume that LEMoE is currently at time $i$ to edit $\text{data}_i$ using module $\text{FFN}_i$. \emph{Left:} When editing data $\text{data}_i$, the prior experts corresponding to previous data are all frozen, leaving only the new model $\text{FFN}_i$ and router trainable. \emph{Right:} In the training stage, depicted by the solid lines, the routing weight $g(i \mid x)$ (abbreviated as $g_i$) is computed using the instance-level embedding and expert key vectors $\{\boldsymbol{k}_1, \boldsymbol{k}_2, \dots, \boldsymbol{k}_i\}$ for expert selection. During inference, as indicated by the dashed lines, the same routing computation is employed to direct the input to the corresponding expert.}
\label{pic:main}
%\vspace{-4mm}
\end{figure*}

\paragraph{Experiments}
To evaluate the model's editing order sensitivity, we employ the same set of editing data and randomly shuffle the order before performing lifelong editing. In this lifelong editing setup, the sequential editing steps is set to 10, with each step editing a batch of 10 (or 100) instances, resulting in a total of 100 (or 1000) edited instances. Each of these two data volumes experiment is conducted 100 times. To assess the relationship between the semantic similarity of the editing inputs and the editing results, we calculate both within-batch semantic similarity ($WBS$) and between-batch semantic similarity ($BBS$) of the editing data. Specially, given dataset $\mathcal{D}_{edit}=\{\mathcal{B}_1,\mathcal{B}_2,\dots,\mathcal{B}_s\}$ with $s$ sequential batches and each batch $\mathcal{B}_i$ contains $n$ edits $\mathcal{B}_i = \{(x_i^e, y_i^e)\}_{i \in [1, n]}$, the $WBS_i$ of $\mathcal{B}_i$ and $BBS$ can be calculated as:
\begin{equation}
WBS_i = \frac{2}{n(n-1)} \sum\nolimits_{1 \leq i < j \leq n}sim(\boldsymbol{e}_i,\boldsymbol{e}_j)  
\end{equation}
%\vspace{-2mm}
\begin{equation}
\label{equ:bgs}
BBS = \frac{2}{s(s-1)} \sum\nolimits_{1 \leq i < j \leq s}sim(\boldsymbol{B}_i,\boldsymbol{B}_j) 
\end{equation}
where $sim(\cdot)$ denotes cosine similarity based on embedding, $\boldsymbol{e}_i = \operatorname{concat}(\operatorname{embed}(x_i^e, y_i^e))$ and $\operatorname{embed}$ represents the embedding vector output from the model's embedding layer. For input consisting of multiple tokens, the embedding is the mean for each token. In Equation \ref{equ:bgs}, $\boldsymbol{B}_i=\mathbb{E}_i(\boldsymbol{e}_i)$ denotes the average semantic vector for the $i$-th group. We employ LLaMA2-7B and ZsRE, with all other experimental settings consistent with \S \ref{subsec:catastrophic}.

\paragraph{Results} 
On the left of Figure \ref{pic:violine}, the MoE adaptor demonstrates significant order sensitivity. Varying the order of the same editing data leads to performance fluctuations exceeding 20 points. In 100 data editing, these fluctuations even range from 30 to 90 point. This substantial impact of editing order on model performance suggests that adopting order align with model preferences can greatly enhance editing efficiency.
The results on the right reveal that higher within-batch semantic similarity and lower between-batch semantic similarity correlate with better editing results. These provide insights for designing more effective editing order. 

%\vspace{-2mm}
\section{Methods}
Based on the above insights, in this section, we provide a detailed introduction to LEMoE an advanced MoE adaptor with new module inserting, KV anchor routing and clustering-based order planning.

%\vspace{-2mm}
\subsection{New Module Inserting}
\label{subsec:New Module Inserting}
Inspired by~\cite{memoe}, LEMoE introduces multiple parallel experts within the transformer feed-forward network (FFN) via a bypass mechanism, while freezing all the model's original parameters. This module is applied in only one transformer block of the entire model. The choice to use the FFN module is motivated not only by its traditional role in MoE but also by recent experimental findings from knowledge probing technologies that suggest the MLP layers within the FFN store knowledge~\cite{DBLP:conf/emnlp/GevaSBL21, rome, MEMIT}. The bypass mechanism preserves all the original parameters of the model, enhancing the locality of model editing.

However, in conventional MoE adaptor, all experts are sequentially trained without any mechanism to protect prior editing knowledge, which allows current edits to easily affect previous ones and leads to severe catastrophic forgetting. Meanwhile, experimental results indicate that a single FFN expert is sufficient to learn a batch of editing data~\cite{memoe}. Therefore, in LEMoE, we adopt a straightforward method to maintain edits from previous learning phases. As shown on the left of Figure \ref{pic:main}, when facing a new batch of editing data in the sequence, we add a new FFN module as an expert to learn this batch of data and freeze the expert network corresponding to the previous data. By aligning the expert networks in MoE architecture with the data batches in lifelong editing, we mitigate the adverse effects of current data editing on previous edits, thereby alleviating catastrophic forgetting from a model mechanism perspective.

Specially, when the $t+1$\textsuperscript{th} batch in lifelong editing dataset arrives, the LEMoE adaptor integrates previous $t$ experts denoted as $f_1,f_2,\dots,f_t$, a router $g(i\mid x)$ which outputs the corresponding coefficients for each $f_i$ based on the input $x$ along with a newly added expert $f_{t+1}$. The output $h$ of this module can be expressed as:
%The output $h$ of the forward process of this module can be expressed as:
%\vspace{-4mm}
\begin{equation}
\begin{aligned}
& h(x) = \mathbf{W}_0 \cdot x + \lambda \sum_{i=1}^{t+1} g(i \mid x) f_i(x) \\
&g(i \mid x) = \operatorname{Top}_k(\frac{e^{r(x)_i}}{\sum e^{r(x)_j}}) \\
\end{aligned}
\label{equ:router}
\end{equation}
where $\mathbf{W}_0$ is the frozen original FFN parameters, $r(x)$ is the routing strategy and is modeled by one MLP in conventional MoE. $\lambda$ is a non-negative weighting coefficient used to balance the old and new knowledge and usually set to 1.

\subsection{KV Anchor Routing}
\label{subsec:KV Anchor Routing}
We propose the KV anchor routing to align the training and inference processes for expert selection, thereby enhancing routing consistency and addressing catastrophic forgetting at routing level.

During the training phase, when the $t$-th batch in the lifelong editing dataset arrives, we freeze the parameters of all previous experts $f_1,f_2,\dots,f_{t-1}$ and introduce a new expert $f_t$ to accomplish the current batch editing. We allocate a key vector $\boldsymbol{k}_i$ for each expert $f_i$ (at time $t$, only $f_t$ is allocated a new key, while the keys corresponding to previous experts remain frozen) and compute instance-level embedding features of the input as values.

The KV anchor process begins with the $j$-th input sentence $\text{X}_{jt}=\{x_i^{jt}\}_{i=1}^L$ of the current $t$-th batch data passing through the embedding layer of the LLM backbone to obtain $\boldsymbol{E}_j^t$ (we omit the superscripts $t$ for simplicity). Since $\boldsymbol{E}_j \in \mathbb{R}^{m \times d}$ and each key vector $\boldsymbol{k}_i \in \mathbb{R}^{d}$ have different sequence lengths, we apply mean-pool operation on the length dimension of $\boldsymbol{E}_j$, and obtain $\boldsymbol{e}_j \in \mathbb{R}^d$. Then $\boldsymbol{e}_j$ is fed to a sub-network to project it into the spaces of the key vectors for better feature alignment. This consists of down and up projection:
%\vspace{-2mm}
\begin{equation}
\boldsymbol{v}_j = \boldsymbol{W}^{\textrm{up}} (\textrm{SiLU}(\boldsymbol{W}^{\textrm{down}} \cdot\boldsymbol{e}_j))
\label{equ:kv1}
\end{equation}
where $\boldsymbol{W}^{\textrm{down}} \in \mathbb{R}^{d_p \times d}$ and $\boldsymbol{W}^{\textrm{up}} \in \mathbb{R}^{d \times d_p}$ are learnable projection parameters. Then, the router $g(i \mid \boldsymbol{e}_j)$ in Equation \ref{equ:router} can be defined as:
%\vspace{-2mm}
\begin{equation}
    g(i \mid \boldsymbol{e}_j) = \operatorname{Top}_k (\frac{\mathrm{e}^{\boldsymbol{k}_j \boldsymbol{v}_j}} {\sum_{i=1}^{t} \mathrm{e}^{\boldsymbol{k}_i \boldsymbol{v}_i}})
\label{equ:kv2}
\end{equation}
The output for the input token $x_i^{jt}$ of aggregated experts can be obtained:
%\vspace{-2mm}
\begin{equation}
\label{equ:kv output}
h(x_i^{jt}) = \mathbf{W}_0 \cdot x + \lambda \sum_{i=1}^{t} g(i \mid \boldsymbol{e}_j) f_i(x_i^{jt}) \\
\end{equation}

During the inference phase, when testing data from different batches arrive, they undergo the same routing computation of Equations \ref{equ:kv1} and \ref{equ:kv2} to reach the appropriate expert. Although there is a possibility that the earlier testing data may be routed to the experts corresponding to the later data, we mitigate this issue by employing the clustering-based editing order selection described in \S \ref{subsec:Clustering-based Order Planning} which reduces semantic similarity between batches. In summary, aligning the routing computations during training and testing phases through KV anchors enhances routing consistency and further mitigates catastrophic forgetting.

\begin{table*}[t]
\caption{Lifelong editing results. \textbf{Bold} is the best result. $T$: Num Edits.}
%\vspace{-2mm}
\centering
% \setstretch{1.2}
\resizebox{\linewidth}{!}{
\begin{tabular}{lcccc|cccc|cc|cc}
    \toprule
    \multirow{3}{*}{\textbf{Method}} & \multicolumn{8}{c|}{\textbf{ZsRE}} & \multicolumn{4}{c}{\textbf{SelfCheckGPT}}\\
    \cmidrule(lr){2-13}
    & \multicolumn{4}{c|}{$T=100$} & \multicolumn{4}{c|}{$T=1000$} & \multicolumn{2}{c|}{$T=100$} & \multicolumn{2}{c}{$T=600$} \\
    \cmidrule(lr){2-13}  & Rel.$\uparrow$ & Gen.$\uparrow$ & Loc.$\uparrow$ & Avg.$\uparrow$ & Rel.$\uparrow$ & Gen.$\uparrow$ & Loc.$\uparrow$ & Avg.$\uparrow$ & PPL$\downarrow$ & Loc.$\uparrow$ & PPL$\downarrow$ & Loc.$\uparrow$  \\
    \midrule
    \multicolumn{13}{c}{\texttt{\textbf{\hspace{12mm} LLaMA2-7B}}} \\
    \midrule
    FT-L    & 0.30 & 0.27 & 0.23 & 0.27   & 0.19 & 0.16 & 0.03 & 0.13    &33.06 &0.41  &69.22   &0.26   \\
    FT-EWC  & 0.83 & \textbf{0.74} & 0.08 & 0.55 & 0.76 & \textbf{0.69} & 0.08 & 0.51    &2.10  &0.16  &4.56    &0.24   \\
    MEND    & 0.00 & 0.00 & 0.00 & 0.00   & 0.00 & 0.00 & 0.00 & 0.00    &10.04 &0.88  &1847.90 &0.00   \\
    ROME    & 0.23 & 0.22 & 0.04 & 0.16   & 0.01 & 0.01 & 0.00 & 0.01    &94.15 &0.05  &104.93  &0.02   \\
    MEMIT   & 0.76 & 0.68 & 0.85 & 0.76   & 0.69 & 0.65 & 0.62 & 0.65    &7.18  &0.96  &13.47   &0.94   \\
    DEFER   & 0.20 & 0.12 & 0.27 & 0.20   & 0.03 & 0.03 & 0.74 & 0.27    &8.91  &0.19  &19.16   &0.12  \\
    GRACE   & \textbf{0.96} & 0.00 & \textbf{1.00} & 0.65 & \textbf{0.97} & 0.08 & \textbf{1.00} & 0.68    & 9.44 & \textbf{1.00} & 9.34 & \textbf{1.00} \\
    MEMoE   &0.72 & 0.46 & \textbf{1.00} & 0.73   &0.70 & 0.43 & \textbf{1.00} & 0.71    &3.00  &\textbf{1.00}  &6.59    &\textbf{1.00}   \\
    \midrule 
    \rowcolor{blue!8} 
    \textbf{LEMoE}  &0.83 &0.62 &\textbf{1.00} &\textbf{0.82}   &0.80 &0.60 &\textbf{1.00} &\textbf{0.80}    &\textbf{2.01} &\textbf{1.00}  &\textbf{3.36} &\textbf{1.00}   \\
    \midrule
    \multicolumn{13}{c}{\texttt{\textbf{\hspace{12mm} Mistral-7B}}} \\
    \midrule
    FT-L       & 0.11 & 0.10 & 0.02 & 0.08 & 0.16 & 0.13 & 0.01 & 0.10     &1594.93  &0.00  &-    &-      \\
    FT-EWC     & 0.82 & \textbf{0.72} & 0.09 & 0.54 & 0.76 & \textbf{0.69} & 0.09 & 0.51     &4.73     &0.17  &5.46 &0.25   \\
    MEND       & 0.00 & 0.00 & 0.00 & 0.00 & 0.00 & 0.00 & 0.00 & 0.00     &23114.94 &0.01  &-    &-      \\
    ROME       & 0.05 & 0.05 & 0.02 & 0.04 & 0.04 & 0.04 & 0.02 & 0.03     &103.75   &0.03  &241.17 &0.01 \\
    MEMIT      & 0.73 & 0.71 & 0.88 & \textbf{0.77} & 0.73 & 0.70 & 0.62 & 0.68     &3.22     &0.97  &7.28  &0.95  \\
    DEFER      & 0.28 & 0.17 & 0.26 & 0.24 & 0.02 & 0.02 & 0.67 & 0.24     &9.54     &0.43  &24.16 &0.13  \\
    GRACE      & \textbf{1.00} & 0.00 & \textbf{1.00} & 0.67 & \textbf{1.00} & 0.02 & \textbf{1.00} & 0.67       &9.53 &\textbf{1.00}  &9.57 &\textbf{1.00}  \\
    MEMoE   & 0.70 & 0.43 & \textbf{1.00} & 0.71   & 0.70 & 0.41 & \textbf{1.00} & 0.70    &4.96  &\textbf{1.00}  &8.91    &\textbf{1.00}   \\
    \midrule  
    \rowcolor{blue!8}
    \textbf{LEMoE} & 0.78 & 0.52 &\textbf{1.00} & \textbf{0.77}   &0.75 &0.48 &\textbf{1.00} & \textbf{0.74}    &\textbf{3.03} &\textbf{1.00}  &\textbf{4.39} &\textbf{1.00} \\
    \bottomrule 
\end{tabular}
}
\label{tab:main_results}
%\vspace{-4mm}
\end{table*}

\subsection{Clustering-based Order Planning}
\label{subsec:Clustering-based Order Planning}
In \S \ref{subsec:Order Sensitivity Analysis}, we observed a correlation between improved editing performance and editing order characterized by high between-batch semantic similarity and low within-batch semantic similarity. 
This suggests that editing performance can be improved by selecting editing order that align with model biases.
Additionally, this objective aligns with the goals of clustering algorithms which aim for high intra-cluster similarity and low inter-cluster similarity. Therefore, we employed the K-means algorithm to group the editing data based on semantic similarity and preferentially selected data from the same cluster for each batch during editing. Experimental results indicate that this straightforward approach is highly effective.

\section{Experiments}
\label{sec:exp}
\subsection{Experimental Setups}
\label{subsec:exp_setups}

\paragraph{Datasets and Metrics}
\label{par:datasts and metrics}
We used two lifelong model editing datasets: ZsRE~\cite{ZsRE} and SelfCheckGPT~\cite{selfcheckgpt}. ZsRE is a context-free Question Answering (QA) dataset built upon zero-shot relation extraction, and we adopt the split provided by \cite{KE_survey}. SelfCheckGPT is a dataset for evaluating the performance of model editing methods on mitigating model hallucination, and we We followed the GRACE~\cite{GRACE} data processing approach. Further details about the datasets are provided in Appendix \ref{apd:dataset_details}. In terms of evaluation metrics, we use the three metrics mentioned in \S \ref{sec:preliminaries}: Reliability (Rel.), Generality (Gen.), and Locality (Loc.), along with the average scores (Avg.) over these metrics. Notably, for the SelfCheckGPT dataset,following~\cite{wise}, we use the perplexity (PPL) to verify Reliability, and there is no proper metric for generalization.

\paragraph{Baselines}
We compare LEMoE with the following four types baselines:
%\vspace{-2mm}
\begin{itemize}[leftmargin=0.5cm]
	\setlength\itemsep{0em}
    \item \textbf{Fine-tuning based methods:} FT-L~\cite{rome}, FT-EWC~\cite{DBLP:journals/corr/KirkpatrickPRVD16}. FT-L directly fine-tunes a single layer’s FFN and FT-EWC is a continual learning fine-tuning methods based on Elastic Weight Consolidation. 
    \item \textbf{Locate and edit methods:} ROME~\cite{rome}, MEMIT~\cite{MEMIT}. These methods treat FFN of transformer as a linear associative memory apply causal tracing to locate the editing area within model.
    \item \textbf{Meta-learning methods:} MEND~\cite{mend}. MEND learns a hyper-network using additional training data to transform gradient obtained by standard fine-tuning. 
    \item \textbf{Memory based methods:} DEFER~\cite{serac}, GRACE~\cite{GRACE}. DEFER is inspired by SERAC~\cite{serac} using an external cache to store explicit editing cases, while GRACE adopts a codebook to store relevant edits.
\end{itemize}

%\vspace{-1mm}
\paragraph{Implementation Details}
We selected LLaMA2-7B and Mistral-7B as base models. The modification was applied to layer 18 with $top_k=1$. Due to limited computational resources, we were able to add a maximum of 5 FFN experts. Consequently, the sequential editing steps were set to 5 and each step contains a batch of 25 (or 200) instances, resulting in a total of 100 (or 1000) editing instances. We use AdamW~\cite{DBLP:conf/iclr/LoshchilovH19} as the optimizer with a learning rate of 2e-4. Further details are provided in the Appendix \ref{apd:implementation_details}.

%\vspace{-1mm}
\subsection{Main Results}
Experimental results are presented in Table \ref{tab:main_results}. 
On ZsRE dataset, LEMoE outperforms all the comparison methods in average scores, achieving up to a 12.68\% improvement over the nearest competitor.
While MEMIT shows comparable performance at $T$=100, our method demonstrates a substantial performance gap in longer sequence editing task. In Locality, our method consistently scores 1.00, indicating minimal impact on irrelevant inputs. Although GRACE and FT-EWC achieve higher score in Reliability and Generality, these methods make great sacrifices in generality and locality respectively. Only our method achieves a better balance.
 
The performance advantage of LEMoE is more pronounced on SelfCheckGPT dataset, maintaining the lowest perplexity scores of 3.36 and 4.39 at $T$ = 600, with a maximum improvement of 26.31\% over the nearest competitor and a constant locality score of 1.00. In summary, across the two datasets and eight baselines, our method shows a clear performance advantage.

\begin{table}[t]
\centering
%\vspace{-2mm}
\caption{Batch editing results. \textbf{Bold} is the best result, and \underline{underline} is the second-best. \texttt{ZsRE. LLaMA2-7B.}}
%\vspace{-2mm}
\fontsize{6}{6}\selectfont
\setlength{\heavyrulewidth}{0.17mm}
\setlength{\lightrulewidth}{0.1mm}
\setlength{\cmidrulewidth}{0.1mm}
\resizebox{1.0\linewidth}{!}{
    \begin{tabular}{lccc|c}
    \toprule
    \multirow{1}{*}{\textbf{Method}}  & Rel.$\uparrow$ & Gen.$\uparrow$ & Loc.$\uparrow$ & Avg.$\uparrow$ \\ 
    \midrule
    FT-L   & 0.14 & 0.13 & 0.70 & 0.32 \\
    MEND   & 0.01 & 0.28 & 0.97 & 0.34   \\
    MEMIT  & 0.24 & 0.40 & 0.17 & 0.27   \\
    SERAC  & 0.89 & 0.16 & 0.81 & 0.62 \\
    GRACE  & 0.95 & 0.38 & \textbf{1.00} & 0.78 \\
    MEMoE  & \textbf{1.00} & \textbf{0.90} & \textbf{1.00} & \textbf{0.97} \\
    \rowcolor{blue!8}
    \textbf{LEMoE} & \textbf{1.00} & \underline{0.88} & \textbf{1.00} & \underline{0.96}  \\ 
    \bottomrule
    \end{tabular}
}
\label{tab:batch editing}
%\vspace{-3mm}
\end{table}

\begin{table}[t]
\centering
%\vspace{-2mm}
\caption{Scaling to 3K edits. \texttt{ZsRE. LLaMA2-7B}.}
%\vspace{-2mm}
\fontsize{14}{20}\selectfont 
% \setstretch{1.2}    
\setlength{\heavyrulewidth}{0.4mm}
\resizebox{1.0\linewidth}{!}{
    \begin{tabular}{lcccc|cccc}
    \toprule
    \multirow{3}{*}{\textbf{Method}} & \multicolumn{4}{c|}{$T = 2000$} & \multicolumn{4}{c}{$T = 3000$} \\
    \cmidrule(lr){2-9}
    & Rel. & Gen. & Loc. &Avg. & Rel. & Gen. & Loc. &Avg. \\
    \midrule
    GRACE &	\textbf{0.96} &	0.03 &	\textbf{1.00} &0.66 &\textbf{0.96} &0.03 &\textbf{1.00} &0.66 \\
    MEMIT &0.64 &\textbf{0.58}  &0.55  &0.59 &0.58 &\textbf{0.53} &0.47 &0.53 \\
    \rowcolor{blue!8}
    LEMoE &\underline{0.74} &\underline{0.50} &\textbf{1.00} &\textbf{0.75} &\underline{0.70} &\underline{0.48} &\textbf{1.00} &\textbf{0.73} \\
    \bottomrule 
    \end{tabular}
}
\label{tab:scaling_3K}
%\vspace{-4mm}
\end{table}

\begin{table}[t]
\centering
%\vspace{-2mm}
\caption{Results of ablation study using 1k edits. \textbf{Bold} is the best result. \texttt{ZsRE. LLaMA2-7B.}}
%\vspace{-2mm}
\fontsize{9pt}{10.8pt}\selectfont 
\setlength{\heavyrulewidth}{0.25mm}
\resizebox{1.0\linewidth}{!}{
    \begin{tabular}{lccc|c}
    \toprule
     & Rel.$\uparrow$ & Gen.$\uparrow$ & Loc.$\uparrow$ & Avg.$\uparrow$ \\ 
    \midrule
    \rowcolor{blue!8}
    LEMoE             & 0.82 & 0.59 &\textbf{1.00} &\textbf{0.80}  \\
    + Conventional Routing     & 0.70 & 0.43 & \textbf{1.00} & 0.71   \\
    + Knowledge Routing  & 0.72 & 0.48 & \textbf{1.00} & 0.73   \\
    + Token-level Embed. & 0.75 & 0.46 & \textbf{1.00} & 0.74 \\
    + Entity-level Embed.& 0.80 & 0.57 & \textbf{1.00} & 0.79 \\
    - Order Planning     & 0.78 & 0.55 & \textbf{1.00} & 0.78  \\
    + Hierarchical Cluster  & 0.82 & 0.58 & \textbf{1.00} & 0.80  \\ 
    \bottomrule
    \end{tabular}
}
\label{tab:ablation}
%\vspace{-4mm}
\end{table}

\section{Detailed Analysis and Discussion}
\subsection{Batch Editing}
\label{subsec:batch editing}
Considering the significant performance advantages of conventional MoE adaptor in batch editing~\cite{memoe}, we aim to evaluate the changes in batch editing performance of its improved version, LEMoE, after applying the proposed optimizations. The batch size is set to 30 here.
As shown in Table \ref{tab:batch editing}, LEMoE continues to excel in batch editing, achieving perfect reliability and locality scores of 1.00, with only a slight decline in generalization. Overall, LEMoE's performance is nearly on par with the original MoE, demonstrating the dual advantages in both batch editing and lifelong editing.

%\vspace{-2mm}
\subsection{Longer Sequence Editing}
We scale the number of lifelong editing to 3K in Table \ref{tab:scaling_3K}. We observe that LEMoE outperforms the strongest baselines MEMIT and GRACE. GRACE excels in reliability but almost entirely loses generalization. While MEMIT demonstrates better generalization, its lower locality scores indicate a significant impact on unrelated data inputs, potentially affecting the model's general ability~\cite{general_ability_paper}. Only our method achieves a balanced editing performance. Moreover, the performance advantage of our approach increases with the number of edits, highlighting the potential of LEMoE to handle extremely long sequential editing.

%\vspace{-2mm}
\subsection{Ablation Study}
\label{subsec:ablation study}
We present a series of ablation studies to evaluate the influence of various model components, including routing strategies, embedding levels and order planning. The experimental results are shown in Table \ref{tab:ablation}. Conventional routing means the router is modeled by an MLP, knowledge (anchor) routing is the routing strategy in MEMoE and entity-level embedding means substitute the embeddings of named entities from the input for $e_j$ in Equation \ref{equ:kv1}. More details in Appendix \ref{apd:Implementation for Ablation Study}.

We observe that: 
(1) Different model settings exhibit minimal impact on locality but significantly affect generality. 
(2) Alteration in routing strategy notably affect reliability and generality, and conventional routing yields the lowest scores across all metrics.  Meanwhile, employing knowledge routing marginally enhances performance yet still lags behind LEMoE, highlighting the pronounced efficacy of KV-anchor routing.
(3) Using token-level embeddings for routing inputs notably diminishes model generality. A possible reason is that token representation may not be suitable for measuring semantic similarity in autoregressive LLMs~\cite{wise}, thereby hindering router's ability to router the same input to the same expert.
(4) Substituting hierarchical clustering for K-means in editing order planning minimally impacts model performance, yet K-means demonstrates higher computational efficiency. This may stem from our utilization of a small number of clusters and a large batch size during dataset construction, which provides clustering algorithms with greater fault tolerance, thereby partially masking the performance differences between the two clustering algorithms.

%\vspace{-2mm}
\section{Conclusion}
In this paper, We propose LEMoE, an advanced MoE adaptor for lifelong model editing. We analyze three factors influencing the effectiveness of MoE adaptor in lifelong model editing. Then, we propose three optimization modules. These modules align the routing computation processes between training and testing phases, ensuring the same inputs are routed to the same experts. Experimental results validate the effectiveness of LEMoE across multiple models and datasets.

\section*{Ethics Statement}
Our research on model editing and the proposed LEMoE module adheres to the ethical guidelines outlined by the ACL Ethics Policy. The primary objective of our work is to enhance lifelong editing performance in LLMs. 
We recognize the critical importance of addressing privacy concerns when model editing publicly accessible, centralized LLMs with private data. And, we acknowledge the potential risks associated with direct parameter edits within models, especially when using harmful data, which require careful mitigation. It’s essential to bear in mind that ill-intentioned model editing could lead the model to generate harmful or inappropriate outputs. Therefore, ensuring safe and responsible practices in model editing is of paramount importance. The application of these techniques should be guided by ethical considerations, with safeguards in place to prevent misuse and the production of harmful results. 
Our commitment to accountability, responsible governance, and continuous ethical assessment underscores our dedication to upholding the highest standards of integrity in the development and deployment of model editing methods.

\section*{Limitations}
There are several limitations to consider for future directions of model editing of large language models. 
Firstly, when the learning sequence scales to more data, such as hundreds of batches or tens of thousands of editing instances, continually allocating an expert block for each batch would lead to significant computational and storage costs. Therefore, exploring methods to prune and merge similar experts in the continual learning process presents an interesting research direction.
Secondly, our work primarily focuses on the acquisition of factual knowledge, neglecting other types of knowledge. We prioritize the accuracy of knowledge learning while paying less attention to other aspects, such as knowledge reasoning abilities.
Thirdly, due to hardware constraints, our investigation was limited to models with up to 7 billion parameters with 5 experts. Additionally, we concentrated on decoder-only autoregressive models, excluding encoder-decoder architectures. Further research that replicates our study using larger-scale models with much more experts and different architecture would be beneficial in confirming our findings.

\section*{Acknowledgements}
This research is supported by the National Natural Science Foundation of China (No.62106105), the CCF-Baidu Open Fund (No.CCF-Baidu202307), the Scientific Research Starting Foundation of Nanjing University of Aeronautics and Astronautics (No.YQR21022), and the High Performance Computing Platform of Nanjing University of Aeronautics and Astronautics.

\bibliography{custom}

\appendix
\input{apd_related_work}
\input{apd_implementation_details}

\section{More Results and Analyses}

\newcommand{\greenyes}{\textcolor{green}{\ding{51}}}
\newcommand{\bluehalf}{\textcolor{cyan}{\ding{52}\rotatebox[origin=c]{-9.2}{\kern-0.7em\ding{55}}}}
\newcommand{\redno}{\textcolor{red}{\ding{55}}}
\begin{table*}[t]
    % \color{blue}
    \small
    \centering
    \caption{Failure cases of LEMoE. \bluehalf represents errors in part of the tokens, \redno represents complete output errors (i.e.,factual failures), and \greenyes indicates the expected exact match. Italics correspond to generality prompt. \texttt{LLaMA2-7B.}}
    % \begin{tabular}{l@{\hspace{2pt}}ccc}
    \begin{tabular}{l@{\hspace{2pt}}>{\raggedright\arraybackslash}p{0.48\textwidth}>{\raggedright\arraybackslash}p{0.18\textwidth}>{\raggedright\arraybackslash}p{0.23\textwidth}}
        \toprule
        & \normalsize Prompt & \normalsize Edit Target & \normalsize Post-Edit Output \\
        \midrule
        \multirow{4}{*}{$i$}  & What level is Javan surili's iucn conservation status? & critically threatened & near threatened  \bluehalf \\
         & \textit{What is Javan surilis ucn conservation status?} & critically threatened & threatened  \bluehalf \\
         & The point in time of Air France Flight 447 was when? & 12 July 1944 & 12 July 1967 \bluehalf \\
         & \textit{When did Air France Flight 447 occur?} & 12 July 1944 & 12 July 1967 \bluehalf \\
        \midrule
        \multirow{4}{*}{$ii$} &Which war was William Babcock Hazen in?  & World War II & US Civil War  \redno \\
        &\textit{What war did William Babcock Hazen go to?}  & World War II & Spanish Civil War  \redno \\
        & When was the inception of Parcelforce? & 1961 & 1963  \redno \\
         & \textit{When was Parcelforce formed?} & 1961 & 1960  \redno \\
        \midrule
        \multirow{2}{*}{$iii$} &What team is Nicolas Raffault associated with? & Arizona Coyotes & Aqua  \redno \\
         & \textit{Which team is Nicolas Raffault associated with?} & Arizona Coyotes & Arizona Coyotes  \greenyes \\
         &What sports team was Petteri Nummelin a member of? & Columbus Blue Bombers & Cleveland Monsters  \redno \\
         & \textit{In which sports team was Petteri Nummelin a member?} & Columbus Blue Bombers & Columbus Blue Bombers  \greenyes \\
        \midrule
        \multirow{4}{*}{$iv$} & What level is Javan surili's iucn conservation status? & critically threatened & nearlly threatened  \greenyes \\
         & \textit{What state is Qaleh Lan in?} & critically threatened & unknown  \redno \\
         & When did Battle of the Java Sea occur? & 27 February 1942 & 27 February 1942 \greenyes \\
         & \textit{When did the battle on the Java Sea begin?} & 27 February 1942 & 1942 \bluehalf \\
        \bottomrule
    \end{tabular}
\label{tab:case study}
\end{table*}

\subsection{More results for Influencing Factors}
\label{apd:more_influencing_factors}
In \S \ref{subsec:catastrophic}, we employed two different evaluation methods: (1) a standard evaluation conducted on all edited data only after all edits were completed, and (2) an evaluation conducted immediately after each edit to assess the effectiveness of these edits at the current stage. Figure \ref{pic:factors12} shows the variations in the reliability metrics, and we further provide the changes in all three metrics here. To better illustrate these trends, we averaged the metrics over every four steps in a sequence of 100 editing steps. As shown in Figure \ref{pic:ablation_3metric}, both reliability and generalization exhibit catastrophic forgetting phenomenon, where subsequent edits significantly affect the performance on prior data. This effect is most pronounced in the reliability metric. Additionally, around step 80, minimal fluctuations in the current editing reliability result in substantial oscillations in generality. This can be attributed to the fact that, like human, a model must first accurately learn knowledge before it can generalize that knowledge. Thus, the generality metric is, to some extent, contingent upon reliability. Regarding locality, the overall level remains consistently high, above 0.97, and thus the graph shows no discernible pattern of fluctuations. This further corroborates that knowledge editing through bypass mechanisms minimally impacts the model's generalization capability \cite{memoe}.

\subsection{Case Study}
In Table \ref{tab:case study}, we present bad cases of using LEMoE to edit the \texttt{LLaMA-2-7B} on the ZsRE dataset and mitigating these failures is critical for future work in model editing. We observe that:

$i)$ errors occur only in part of the tokens, and these errors constitute a large proportion of the bad cases, indicating that the edits have not been sufficiently fitted. We wonder whether employing different learning rates and epochs for each batch in lifelong editing could alleviate this issue through more refined training.

$ii)$ displays cases where the entire output is incorrect. These types of errors are the most common occurrences.

$iv)$ presents cases of generalization failure. For example in prompt of last line, where the model answered ``1942'' which is partially correct, but did not fully follow the ground truth, indicating significant room for improvement in the accuracy of generalized edits.

\begin{table}[t]
\centering
\tiny
\caption{Performance of LEMoE on different layer of \texttt{LLaMA2-7B} using \texttt{ZsRE.}}
\resizebox{1.0\linewidth}{!}{
\begin{tabular}{lccc|c}
\toprule
Layer & Rel.$\uparrow$ & Gen.$\uparrow$ & Loc.$\uparrow$ & Avg.$\uparrow$ \\
\midrule
0 & 0.28 & 0.14 & 1.00 & 0.47 \\
2 & 0.55 & 0.41 & 1.00 & 0.65 \\
4 & 0.52 & 0.34 & 1.00 & 0.62 \\
6 & 0.44 & 0.25 & 1.00 & 0.56 \\
8 & 0.48 & 0.26 & 1.00 & 0.58 \\
10 & 0.47 & 0.25 & 1.00 & 0.57 \\
12 & 0.54 & 0.26 & 1.00 & 0.60 \\
14 & 0.58 & 0.33 & 1.00 & 0.64 \\
16 & 0.77 & 0.55 & 1.00 & 0.77 \\
\rowcolor{blue!8}
18 & \textbf{0.80} & \textbf{0.60} & \textbf{1.00} & \textbf{0.80} \\
20 & 0.76 & 0.59 & 1.00 & 0.79 \\
22 & 0.70 & 0.54 & 1.00 & 0.75 \\
24 & 0.74 & 0.51 & 1.00 & 0.75 \\
26 & 0.77 & 0.56 & 1.00 & 0.78 \\
28 & 0.73 & 0.48 & 1.00 & 0.74 \\
30 & 0.43 & 0.24 & 1.00 & 0.56  \\
\bottomrule
\end{tabular}
}
%\vspace{-3mm}
\label{tab:ablation layer}
\end{table}

Meanwhile, in $iii)$ we surprisingly find that even when LEMoE errs on the Edit Prompt, it can correctly answer its paraphrase prompt. Upon closely examining these anomalous cases, we found that they predominantly pertain to question-answering scenarios within sports contexts, such as inquiries about a person's team affiliation. We hypothesize that this phenomenon may stem from the relatively limited number of teams in sports contexts, combined with the higher number of athletes and the occurrence of name duplication. Consequently, the model may accidentally provide correct answers to some of these questions.

In summary, LEMoE can handle contextual information correctly in some cases but falls short in specific editing instructions, suggesting that optimizing editing instructions (modifying the editing context) may be a direction for improvement.

\subsection{More Ablation Results}
As an extension \S \ref{subsec:ablation study}, we evaluate the effectiveness of LEMoE applied to different layers. For experimental setup, we utilize the LLama2-7b model and the ZsRE dataset. Lifelong editing involves 1000 instances, with all the other training hyperparameters consistent as detailed in Appendix \ref{apd:train_details}. Experimental results are depicted in Table \ref{tab:ablation layer}. Notably, the 18-th layer exhibits the most significant editing improvements, achieving peak performance across all metrics. Conversely, the first layer demonstrates the least improvement, and the editing hardly takes effect in the low-level transformer block. In contrast, high-level transformer blocks display pronounced editing effects, maintaining high reliability consistently from the 16-th layer onwards. However, significant degradation in editing efficacy is noted towards the 30-th layer, possibly due to the increased proximity to the output. On the other hand, locality remains unaffected, consistently scoring 1.00. 
Thus, our findings further validate that the high-level transformer blocks of LM based on the transformer architecture contain factual information, and editing of these layers will have a significant effect \cite{SCEN}.

\subsection{LEMoE with LoRA structure}
In the era of LLMs, parameter-efficient fine-tuning (PEFT) methods such as LoRA have proven highly effective and convenient for achieving impressive results across various downstream tasks. LoRA \cite{DBLP:conf/iclr/HuSWALWWC22}, proposes a technique that decomposes the update gradient matrix into two small rank-n matrices, significantly reducing the memory requirements for training LLMs. 
Meanwhile, in fields of MoE, some studies have explored replacing traditional MoE structures with LoRA \cite{DBLP:journals/corr/abs-2309-05444, DBLP:journals/corr/abs-2404-13628}. Consequently, we replace the MLP-based expert networks in LEMoE with LoRA modules. Given the challenging nature of lifelong learning tasks, we evaluate the performance of this low-parameter model structure on batch editing tasks with batch size set to 30.

We investigated the effects of varying the number of experts (Exp.), different LoRA ranks, and different $\text{top}_k$ values. Detailed experimental results are provided in the Table \ref{tab:lora} to facilitate further research. We conducted experiments on all even-numbered layers, expert number in [1,10,20], $\text{top}_k$ in [1,10,20] and LoRA Rank in [2,4,8,16,32,64,128,256,512,1024,2048]. We filter out results with Reliability below 0.1, Generality below 0.1, and Locality below 0.5. It is evident that this method performs poorly in editing the low-level transformer blocks (with results falling below the selection criteria and many being zero, hence not presented in the table). Meanwhile, the higher the layer being edited, the better the performance observed. This LEMoE-LoRA achieved optimal performance with 30 layers, 10 experts, a LoRA rank of 2048, and a $\text{top}_k$ value of 10. 

\input{lora}

\end{document}

%% file: apd_related_work.tex
\section{Related Work}
\subsection{Model Editing}
Model editing is a new and active research area where the goal is to make targeted changes to a pre-trained model’s behavior \cite{KE_survey}. Given the fast-growing parameter sizes of LLMs, frequently updating LLMs with new knowledge through retraining is more and more expensive. Hence, it is vital to effectively edit the LLMs’ knowledge without retraining. Previous studies have explored multiple methods for editing the knowledge of LLMs, which can be broadly categorized into two streams based on whether it alters the parameters of the original model \cite{KE_oppotunity, KE_survey}:

\paragraph{Preserve model parameters:}
(1) \textbf{Retrieve augmentation}. These techniques leverage an external knowledge base to enrich or correct information accessible to language models. These augmented knowledge bases seamlessly integrate with the base model, enabling effective retrieval of relevant information when prompted \cite{MurtyMLR22,MadaanTCY22,LiRZWLVYK23}. For example, IKE \cite{ZhengLDFWXC23} employs an in-context learning approach that adjusts language model outputs using corpus-based demonstrations guided by similarity metrics, thereby obviating the need for gradient-based adjustments. 
(2) \textbf{Adding additional parameters:}
This paradigm involves introducing additional trainable parameters to augment a language model's existing knowledge, while preserving its original parameters in a frozen state. T-Patcher \cite{HuangSZZR023} and CaliNET \cite{DongDSXSL22} exemplify this paradigm by integrating specific neurons or patches into the final layer of their Feed-Forward Networks. T-Patcher assigns individual neurons to each distinct error, while CaliNET incorporates multiple neurons to handle various knowledge scenarios. In contrast, GRACE \cite{GRACE} employs a discrete codebook mechanism to dynamically add and update elements, enhancing the model's predictive capabilities over time. 
(3) \textbf{Meta learning}
Recent meta-learning methods use hypernetworks for aiding editing.
MEND \cite{mend} introduces a hypernetwork designed to decouple fine-tuning gradients into updates that generalize edits without compromising performance on unrelated inputs. To mitigate the cancellation issue inherent in MEND, MALMEN \cite{DBLP:journals/corr/abs-2311-04661} employs a hyper-network to generate weight shifts for editing, formulating the aggregation of these shifts as a least squares problem.

\paragraph{Modify model parameters:}
This methodology begins by identifying parameters associated with specific knowledge and directly adjusting them. The Knowledge Neuron (KN) approach \cite{DBLP:conf/acl/DaiDHSCW22} introduces a technique to attribute knowledge to individual "knowledge neurons" and subsequently updates these neurons accordingly. ROME \cite{rome} utilizes causal mediation analysis to pinpoint areas requiring modification. Both KN and ROME operate under the constraint of editing one factual association at a time. To overcome this limitation, MEMIT \cite{MEMIT} extends ROME's framework, enabling simultaneous editing across multiple instances. Building on MEMIT, PMET \cite{abs-2308-08742} integrates attention values to achieve superior performance enhancements. COMEBA-HK \cite{DBLP:journals/corr/abs-2403-05330} identifies the Local Editing Scope and extends MEMIT for sequential editing.

\subsection{Mixture of Experts}
The concept of MoE, particularly when combined with sparse routing, is recognized for significantly enhancing model capacity with minimal computational overhead \cite{DBLP:journals/jmlr/FedusZS22}. Key distinctions in this approach include: i) adapter experts are not trained during the pre-training of the base model, ii) they are parameter-efficient, and iii) they are tailored to specific tasks, unlike token-level opaque computation units whose specialization is not easily interpretable \cite{DBLP:journals/corr/abs-2401-04088}. Regarding the second point, \cite{DBLP:conf/emnlp/WangAM00AG22, DBLP:journals/corr/abs-2309-05444} utilize routing each example to a set of experts, demonstrating improved performance on unseen tasks. \cite{DBLP:journals/corr/abs-2204-07689} implements a separate router for each task and selects a router from a similar task based on domain knowledge. \cite{DBLP:conf/emnlp/YeZR22} proposes task-level MoEs, where a collection of transformer layers acts as experts, and a router dynamically selects from these experts. Additionally, several recent studies have proposed methods for routing queries to specialized pretrained open-source LLMs \cite{DBLP:journals/corr/abs-2311-08692,DBLP:journals/corr/abs-2309-15789}.

\subsection{Continual Learning}
Continual Learning (CL) \cite{DBLP:journals/corr/abs-2404-16789,DBLP:journals/corr/abs-2402-01364} is an essential aspect of machine learning as it enables models to adapt to new tasks while retaining performance on previous ones. It mainly focus on the issue of catastrophic forgetting in deep learning models when exposed to new knowledge \cite{DBLP:journals/pami/LangeAMPJLST22}. Recent research has explored diverse approaches in this domain. Among these approaches, continual fine-tuning stands out, involving the iterative refinement of LLMs with incoming instances. For instance, \cite{DBLP:conf/acl/LinWLJXRY22} conducts an extensive investigation into this method. However, it has been noted that integrating regularized fine-tuning techniques such as Elastic Weight Consolidation \cite{DBLP:journals/corr/KirkpatrickPRVD16}, Experience Replay \cite{DBLP:conf/nips/RolnickASLW19}, and Maximally Interfered Replay \cite{DBLP:conf/nips/AljundiBTCCLP19} can lead to a decline in performance on earlier tasks while preserving some memory of past inputs. This observation underscores the challenges unique to editing in contrast to conventional continual fine-tuning \cite{DBLP:conf/miccai/HennSJYATSISLT21}, particularly given the uneven distribution of edits. One promising avenue in continual learning involves the adoption of key-value methodologies, inspired by advancements in computer vision \cite{DBLP:conf/nips/LiuWHZMG21,DBLP:conf/nips/OordVK17}. Notably, discrete key-value methods have proven effective in managing shifting distributions \cite{DBLP:conf/icml/TraubleGRMKBS23}. These methods cache values to ensure inputs remain within distribution bounds for downstream encoders, thereby enabling the integration of longer-term memory, contingent on available computational resources.

\subsection{Data Clustering for LLMs}
Data clustering methods for LLMs have been proposed to enhance performance and reduce task interference \cite{DBLP:conf/nips/FiftyAZYAF21,DBLP:journals/corr/abs-2303-14177,DBLP:journals/corr/abs-2312-12379}. These methods include clustering based on similarities computed using tf-idf and neural embeddings, K-means clustering with balanced linear assignment, and soft clustering with Gaussian Mixture Models (GMMs) \cite{DBLP:conf/eacl/ChronopoulouPFD23,DBLP:journals/corr/abs-2303-14177, DBLP:conf/acl/Duan00W0Z20}. Recent work by \cite{DBLP:conf/emnlp/ZhouXM22} highlights the potential of adapter parameters as effective task embeddings for clustering. Additionally, a similar observation regarding task gradients has been made by \cite{DBLP:conf/emnlp/VuWMSTMMI20}.

%% file: apd_implementation_details.tex
\section{Implementation Details}
\label{apd:implementation_details}

\begin{table*}
    \centering
    \caption{An editing dataset example from ZsRE and SelfCheckGPT.}
    \resizebox{\linewidth}{!}{
    \begin{tabular}{ll@{\hspace{8pt}}p{13cm}}
    \toprule
    Dataset & Type & Text \\ 
    % \cmidrule(lr){2-4}
    % & Dataset & N & Pre-edit & Dataset \\
    \midrule
    \multirow{3}{*}{ZsRE} & $\mathbf{x}_i^e, \mathbf{y}_i^e$ & Which continent is Berkner Island in? \textbf{South America} \\
    & $\mathbf{x}_{\text{loc}_i}, \mathbf{y}_{\text{loc}}$ & who gets the golden boot if its a tie? \textbf{shared} \\
    & $\mathbf{x}_{\text{gen}_i}, \mathbf{y}_i^e$ &  On which continent is Berkner Island located? \textbf{South America} \\
    \cmidrule{2-3}
    \multirow{8}{*}{SelfCheckGPT} & $\mathbf{x}_i^e, \mathbf{y}_i^e$ & This is a Wikipedia passage about heinz christian pander. Heinz Christian Pander (1794 - 1865) was a German anatomist and embryologist who was born in Riga, Latvia. He studied medicine at the University of Dorpat and later at the University of Berlin. \textbf{In 1820, he took part in a scientific expedition to Bokhara as a naturalist.} \\
    & $\mathbf{x}_{\text{loc}}, \mathbf{y}_{\text{loc}}$ & Tired and restlessly, drifting in and out of sleep. Hearing crashing and banging, thinking the roof will cave in. Not alert enough to quite know what. \textbf{it was, I yelled loudly for whoever was making those noises at such an hour to stop. They heard and listened, I'm guessing} \\
    \bottomrule\\    
    \end{tabular}
    }
    \vspace{-2mm}
\label{tab:datasets_text}
\end{table*}
\begin{table}
    \centering
    \caption{Dataset statistics for main results.
    \textit{Locality Data} is the irrelevant data of the editing process. $T$ is the number of samples. \textit{Pre-edit} is the unedited model's performance on each dataset.}
    \vspace{-0.2cm}
    \resizebox{\linewidth}{!}{
    \begin{tabular}{cccc}
    \toprule
    \textsc{Setting} & \textsc{Editing Data} & $T$ & Pre-edit (LLaMA/Mistral) \\ 
    % \cmidrule(lr){2-4}
    % & Dataset & N & Pre-edit & Dataset \\
    \midrule
    QA & ZsRE & 1,000 & 0.36/0.39 ACC \\
    Hallucination & SelfCheckGPT & 600 & 27.4/19.4 PPL\\
    \bottomrule\\    
    \end{tabular}
    }
    \label{tab:datasets_number}
    \vspace{-4mm}
\end{table}

\subsection{Datasets Details}
\label{apd:dataset_details}

\paragraph{ZsRE} The ZsRE dataset is a context-free Question Answering (QA) dataset that has been extensively studied in the model editing literature \cite{rome, MEMIT, serac, GRACE, memoe}. Each record in this dataset includes an editing statement $\mathbf{x}_i^e$ with target answer $\mathbf{y}_i^e$, a paraphrase prompt $\mathbf{x}_{\text{gen}_i}$ and and a locality prompt $\mathbf{x}_{\text{loc}}$. We adopt the same train/test split as \cite{mend}, consisting of 163,196 training examples and 19,086 test examples. Notably, MEND is the only method that requires fitting a hyper network on the training set; other methods discard the training set and directly perform edits and evaluations on the test set. For our experiments, we randomly sampled 1k and 3k records from the test set to form the edit sets.

\paragraph{SelfCheckGPT} 
We employ SelfCheckGPT \cite{selfcheckgpt},the same dataset as GRACE, to evaluate the effectiveness of Model Editors in reducing hallucinations in autoregressive language models. This dataset consists of highly inaccurate sentences generated from GPT-3 \cite{DBLP:conf/nips/BrownMRSKDNSSAA20}, which are then replaced with corresponding accurate sentences from Wikipedia. This setup mirrors real-world deployment scenarios where models exhibit "unexpected behaviors". The edits in this dataset are significantly longer compared to ZsRE, presenting a more challenging editing environment. Unlike GRACE, which utilized GPT2-XL (1.5B), our primary experiments use larger LLMs, specifically LLaMA and Mistral, each with 7B parameters. We measure the retention of $\mathbf{x}_{\text{loc}}$ from the base model, RedPajama \cite{redpajama}, a publicly available version of LLaMA’s pre-training data.

\subsection{Implementation of Baselines}
\label{apd:baselines} 
\paragraph{FT-L}
We followed the procedures outlined in \cite{wise}: all other layers of the LLMs remain frozen, and only a single MLP layer undergoes fine-tuning using an autoregressive loss function. Furthermore, we impose a $\text{L}_{\infty}$ norm constraint to ensure that the parameters do not deviate significantly from the pretrained distribution. Employ the Adam optimizer with consideration of learning rates at 1e-5, 1e-4, and 5e-4, and conduct gradient descents for 50 iterations, ultimately reporting the best results at a learning rate of 5e-4.

\paragraph{FT-EWC}
Elastic Weight Consolidation (EWC) effectively mitigates catastrophic forgetting by updating model weights using the Fisher information matrix, which is computed based on past parameter updates and scaled by a factor $\lambda$ \cite{DBLP:journals/corr/KirkpatrickPRVD16}. In line with \cite{GRACE}, our implementation does not incorporate $\text{L}_{\infty}$ norm constraints, setting the learning rate at 1e-2, the $\lambda_ewc$ penalty factor at 0.1, and the number of replay instances at 10.

\paragraph{MEND}
MEND \cite{mend} performs model editing by employing a hyper-network to transform the gradients derived from standard fine-tuning. This process involves decomposing the model gradients into a low-rank format (rank=1) before converting them into new gradients, which are subsequently applied to the target layer for parameter updates. During training, a small auxiliary hyper-network processes editing examples $(\mathbf{x}_i^e, \mathbf{y}_i^e)$ and $(\mathbf{x}_{\text{gen}_i}, \mathbf{y}_i^e)$. The training loss for MEND consists of the standard autoregressive loss combined with the KL divergence loss, measuring the model's output on $(\mathbf{x}_{\text{gen}_i}, \mathbf{y}_i^e)$ before and after editing. This hyper-network is pivotal in the editing procedure.
Due to the substantial computational resources required to train the meta-network for, the results are from \cite{wise}.

\paragraph{ROME}
ROME \cite{rome} employs causal analysis to identify knowledge residing in specific MLP layers and refines the entire matrix via least squares approximation. This approach assumes MLP as the central repository of knowledge \cite{DBLP:conf/emnlp/GevaSBL21}, incrementally injecting individual pieces of information into the MLP through a Lagrangian residual term at each iteration. Following \cite{wise}, in LLaMA and Mistral, ROME edits the fifth layer, while MEMIT edits layers [4,5,6,7,8].

\paragraph{MEMIT}
%MEMIT is batch 版本
The MEMIT utilized in this study, denoted as MEMIT-MASS as introduced in \cite{wise}, differs notably from its original counterpart. In contrast to sequential editing, MEMIT-MASS facilitates batch processing for modifying multiple knowledge fragments concurrently.
Suppose we collect streaming errors as $(\mathcal{X}, \mathcal{Y}) = \{(\mathbf{x}_0, \mathbf{y}_0), (\mathbf{x}_1, \mathbf{y}_1), ..., (\mathbf{x}_T, \mathbf{y}_T)\}$ and inject them collectively into the MLP, it only involves a single editing operation on the original model as $f_{\Theta_{T}} = {\rm MEMIT}(f_{\Theta_{0}}, \mathcal{X}, \mathcal{Y}).$
Despite its drawback of lacking real-time correction capabilities, we include this approach as a baseline in our experimental evaluations, given the extremely bad performance of the original MEMIT framework.

\paragraph{DEFER}
In GRACE, a reimplementation of SERAC \cite{serac} is utilized, denoted as DEFER. DEFER integrates a network denoted as $g$ (akin to the scope classifier in SERAC). This network $g$ predicts whether to rely on: 1) predictions from the LLMs, or 2) predictions from a newly introduced model. This new model, configured as a single-layer linear network $o$ with a sigmoid activation function, parallels the counterfactual model in SERAC. Throughout the editing phase, $g$ and $o$ undergo joint fine-tuning processes. The experiment with learning rates of 7e-5, 7e-4, and 1e-3, and ultimately report using 7e-5 (optimal).

\paragraph{GRACE}
GRACE \cite{GRACE} utilizes a discrete key-value codebook and maintains the codebook throughout the editing flow by adding, expanding, and splitting KEYs. During the inference phase, it retrieves the nearest KEY and determines whether to replace the activation of the hidden layer output. We adhere to the meticulously crafted parameters outlined in the original study, configuring the optimization of the learning rate to a value of 1 and using ``replace last'' to only replace the activation of the last token in autoregressive scenarios.. The iterative process for optimizing these values spans 100 cycles, with an initial $\epsilon=1$.

\begin{figure*}
\centering
\includegraphics[width=\linewidth]{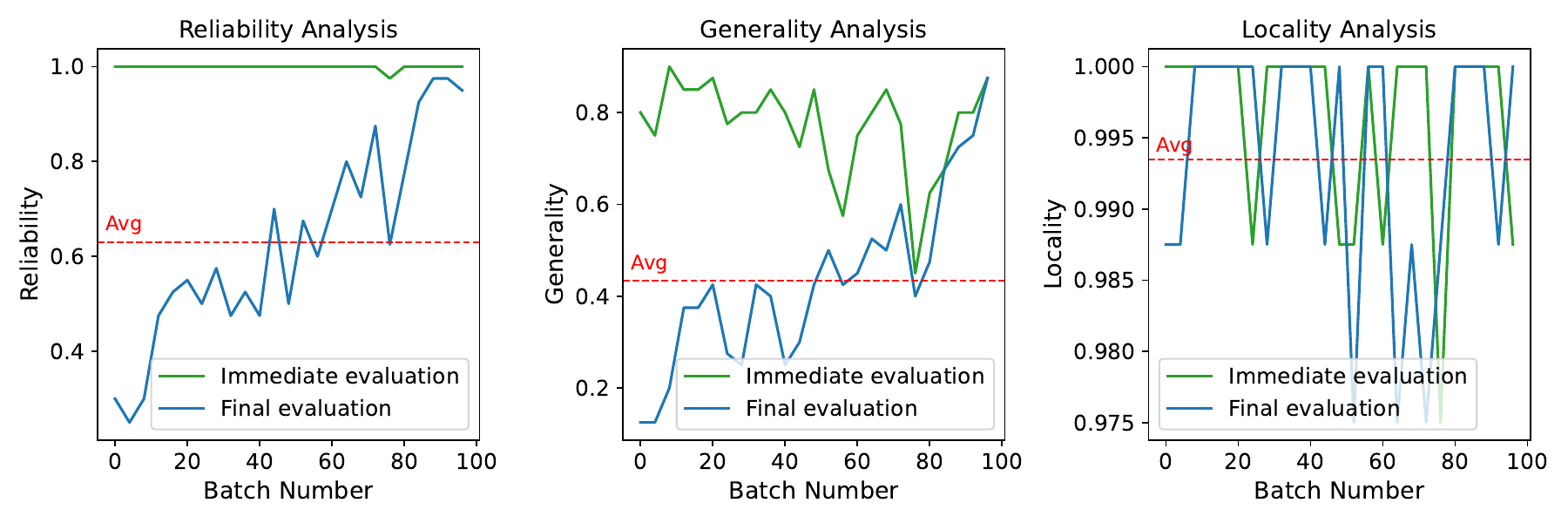}
\caption{Reliability, Generality and Locality of conventional MoE under different stage evaluation. ``Immediate evaluation'' occurs immediately after each edit, ``Final evaluation'' occurs after all edits in lifelong editing. \texttt{Model: LLaMA2-7B. Dataset: ZsRE.}}
\label{pic:ablation_3metric}
\end{figure*}

\paragraph{MEMoE}
MEMoE \cite{memoe} updates knowledge using a bypass MoE structure, keeping the original parameters unchanged to preserve the general ability of LLMs. And, the knowledge anchor routing ensures that inputs requiring similar knowledge are routed to the same expert, thereby enhancing the generalization of the updated knowledge. Following the parameters identified in the original paper, we consulted 4 experts, setting the $top-k$ value to 1 and a learning rate of 2e-4. The modification is applied to \texttt{model.layers[16].mlp.up\_proj.weight} and \texttt{model.layers[16].mlp.down\_proj.weight}.
We also adopt auxiliary loss for balancing the top-k selection of routing following \cite{DBLP:journals/jmlr/FedusZS22}.

\subsection{Training Details of LEMoE}
\label{apd:train_details}

The training loss for the attentive learning of the $t$-th batch data $\mathcal{B}_t$ is:
\begin{equation}
L_{\text{task}} =-\!\sum_{(x_t, y_t) \in \mathcal{T}_t} \! \log P\left(y_t \mid x_t; \theta_m, \theta_f, \theta_{\text{proj}}, \theta_k \right)
\end{equation}
\label{loss}where $\theta_m, \theta_f, \theta_{\text{proj}}$ and $\theta_k$ are parameters of the LLM backbone, the experts, the query projection layer and the set of all key vectors, respectively. And only those parameters belongs to the current $t$-th task are trainable, including $\theta_{f_t}$, $\theta_{\text{proj}}$  and $\theta_{k_t}$.

The hyperparameters for the ZsRE and SelfCheckGPT are identical. Specially, We use the AdamW~\cite{DBLP:conf/iclr/LoshchilovH19} as the optimizer with a learning rate of 2e-4. The modification of the model is applied to \texttt{model.layers[18].mlp.up\_proj.weight} and \texttt{model.layers[18].mlp.down\_proj.weight}. All the experiments are deployed on NVIDIA RTX 3090 Tensor Core GPUs, and we use 4 GPUs for training and single GPU for evaluation. For lifelong editing, due to computational constraints, we can accommodate a maximum of 5 experts. Consequently, the batch size in the sequence is determined by the total number of edits and the number of experts. For instance, when there are 100 edits and 5 experts, the batch size is set to 20; whereas with 1000 edits, the batch size scales up to 200.
For bath editing in \S \ref{subsec:batch editing}, the batch size is 30 and all the other parameters are the same as above.

\subsection{Implementation for Ablation Study}
\label{apd:Implementation for Ablation Study}
In \S \ref{subsec:ablation study}, we conducted an ablation study on several modules of LEMoE. Here, we detail the implementation of these ablations. In Table \ref{tab:ablation}, Conventional routing means the router is modeled by a single-layer MLP, with the preservation of the insertion method. Knowledge routing is the knowledge anchor routing in MEMoE for short, also maintains the insertion method. Token-level embedding involves substituting $e_j$ in Equation \ref{equ:kv output}, which means $ g(i \mid \boldsymbol{e}_j) = g(i \mid x_i^{jt}$. For entity-level embedding, we initially utilize the NLTK tool \footnote{\href{https://www.nltk.org/}{https://www.nltk.org}} for extracting named entities from the input instance. In cases where there are multiple named entities present in the input, we utilize the average pooling of the embeddings of these entities. Subsequently, we replace $e_j$ in Equation \ref{equ:kv1} with this embedding vector as the input to the sub-network to obtain the corresponding ``value'' of the input instances. All the other training hyperparameters are the same detailed in Appendix \ref{apd:train_details}.

%% file: lora.tex
\clearpage
\onecolumn
\begin{longtable}{cccccccc}
\caption{Experimental Results of LEMoE with LoRA module.} \\
\label{tab:lora} \\
\hline
\textbf{Layer} & \textbf{Expert Number} & \textbf{Topk} & \textbf{LoRA Rank} & \textbf{Rel. $\uparrow$} & \textbf{Gen. $\uparrow$} & \textbf{Loc. $\uparrow$} & \textbf{Avg. $\uparrow$} \\
\hline
\endfirsthead

\multicolumn{8}{c}%
{{\bfseries Table \thetable\ Continued from previous page}} \\
\hline
\textbf{Layer} & \textbf{Expert Number} & \textbf{Topk} & \textbf{LoRA Rank} & \textbf{Rel. $\uparrow$} & \textbf{Gen. $\uparrow$} & \textbf{Loc. $\uparrow$} & \textbf{Avg. $\uparrow$} \\
\hline
\endhead

\hline \multicolumn{8}{|r|}{{Continued on next page}} \\ \hline
\endfoot

\hline
\endlastfoot
4 & 10 & 10 & 1024 & 0.13 & 0.10 & 0.53 & 0.26 \\
4 & 10 & 10 & 2048 & 0.30 & 0.23 & 0.68 & 0.41 \\
4 & 20 & 5 & 512 & 0.23 & 0.17 & 0.58 & 0.33 \\
4 & 20 & 10 & 1024 & 0.30 & 0.27 & 0.50 & 0.36 \\
6 & 20 & 5 & 1024 & 0.17 & 0.17 & 0.67 & 0.33 \\
8 & 1 & 1 & 1024 & 0.13 & 0.10 & 0.62 & 0.28 \\
8 & 1 & 1 & 2048 & 0.30 & 0.20 & 0.90 & 0.47 \\
8 & 10 & 1 & 512 & 0.17 & 0.17 & 0.65 & 0.33 \\
8 & 10 & 5 & 2048 & 0.40 & 0.13 & 0.67 & 0.40 \\
8 & 10 & 10 & 1024 & 0.13 & 0.10 & 0.62 & 0.28 \\
8 & 10 & 10 & 2048 & 0.23 & 0.10 & 0.78 & 0.37 \\
8 & 20 & 5 & 1024 & 0.10 & 0.10 & 0.77 & 0.32 \\
8 & 20 & 10 & 512 & 0.17 & 0.13 & 0.90 & 0.40 \\
10 & 10 & 5 & 2048 & 0.13 & 0.20 & 0.93 & 0.42 \\
10 & 20 & 5 & 512 & 0.13 & 0.13 & 0.93 & 0.40 \\
10 & 20 & 5 & 1024 & 0.23 & 0.17 & 0.93 & 0.44 \\
10 & 20 & 10 & 1024 & 0.10 & 0.10 & 0.95 & 0.38 \\
12 & 1 & 1 & 2048 & 0.30 & 0.13 & 0.98 & 0.47 \\
12 & 10 & 1 & 1024 & 0.17 & 0.10 & 0.68 & 0.32 \\
12 & 10 & 5 & 1024 & 0.27 & 0.17 & 0.95 & 0.46 \\
12 & 10 & 5 & 2048 & 0.23 & 0.13 & 0.98 & 0.45 \\
12 & 10 & 10 & 1024 & 0.23 & 0.10 & 0.93 & 0.42 \\
12 & 10 & 10 & 2048 & 0.23 & 0.17 & 0.98 & 0.46 \\
12 & 20 & 5 & 512 & 0.20 & 0.10 & 0.98 & 0.43 \\
12 & 20 & 5 & 1024 & 0.37 & 0.23 & 0.97 & 0.52 \\
12 & 20 & 10 & 512 & 0.23 & 0.13 & 0.98 & 0.45 \\
12 & 20 & 10 & 1024 & 0.20 & 0.10 & 0.98 & 0.43 \\
14 & 1 & 1 & 16 & 0.13 & 0.10 & 0.97 & 0.40 \\
14 & 1 & 1 & 512 & 0.17 & 0.10 & 0.98 & 0.42 \\
14 & 1 & 1 & 1024 & 0.30 & 0.10 & 0.98 & 0.46 \\
14 & 1 & 1 & 2048 & 0.33 & 0.17 & 1.00 & 0.50 \\
14 & 10 & 1 & 128 & 0.13 & 0.10 & 0.68 & 0.31 \\
14 & 10 & 1 & 512 & 0.40 & 0.30 & 0.98 & 0.56 \\
14 & 10 & 1 & 1024 & 0.40 & 0.23 & 1.00 & 0.54 \\
14 & 10 & 1 & 2048 & 0.57 & 0.43 & 1.00 & 0.67 \\
14 & 10 & 5 & 512 & 0.27 & 0.17 & 0.98 & 0.47 \\
14 & 10 & 5 & 1024 & 0.23 & 0.20 & 0.98 & 0.47 \\
14 & 10 & 5 & 2048 & 0.50 & 0.43 & 1.00 & 0.64 \\
14 & 10 & 10 & 16 & 0.13 & 0.10 & 0.95 & 0.39 \\
14 & 10 & 10 & 128 & 0.20 & 0.20 & 1.00 & 0.47 \\
14 & 10 & 10 & 512 & 0.23 & 0.20 & 1.00 & 0.48 \\
14 & 10 & 10 & 1024 & 0.27 & 0.13 & 1.00 & 0.47 \\
14 & 10 & 10 & 2048 & 0.40 & 0.17 & 1.00 & 0.52 \\
14 & 20 & 1 & 1024 & 0.37 & 0.37 & 0.88 & 0.54 \\
14 & 20 & 5 & 128 & 0.17 & 0.13 & 0.98 & 0.43 \\
14 & 20 & 5 & 512 & 0.37 & 0.20 & 1.00 & 0.52 \\
14 & 20 & 5 & 1024 & 0.40 & 0.23 & 0.97 & 0.53 \\
14 & 20 & 10 & 128 & 0.10 & 0.10 & 0.98 & 0.39 \\
14 & 20 & 10 & 512 & 0.20 & 0.13 & 1.00 & 0.44 \\
14 & 20 & 10 & 1024 & 0.30 & 0.20 & 1.00 & 0.50 \\
16 & 1 & 1 & 128 & 0.20 & 0.10 & 1.00 & 0.43 \\
16 & 1 & 1 & 512 & 0.37 & 0.30 & 0.98 & 0.55 \\
16 & 1 & 1 & 1024 & 0.60 & 0.43 & 1.00 & 0.68 \\
16 & 1 & 1 & 2048 & 0.60 & 0.43 & 0.98 & 0.67 \\
16 & 10 & 1 & 512 & 0.43 & 0.37 & 0.95 & 0.58 \\
16 & 10 & 1 & 1024 & 0.47 & 0.33 & 0.93 & 0.58 \\
16 & 10 & 1 & 2048 & 0.73 & 0.50 & 0.93 & 0.72 \\
16 & 10 & 5 & 16 & 0.17 & 0.10 & 0.97 & 0.41 \\
16 & 10 & 5 & 128 & 0.17 & 0.13 & 1.00 & 0.43 \\
16 & 10 & 5 & 512 & 0.50 & 0.27 & 1.00 & 0.59 \\
16 & 10 & 5 & 1024 & 0.37 & 0.17 & 1.00 & 0.51 \\
16 & 10 & 5 & 2048 & 0.53 & 0.20 & 1.00 & 0.58 \\
16 & 10 & 10 & 16 & 0.20 & 0.10 & 1.00 & 0.43 \\
16 & 10 & 10 & 128 & 0.33 & 0.27 & 1.00 & 0.53 \\
16 & 10 & 10 & 512 & 0.40 & 0.33 & 1.00 & 0.58 \\
16 & 10 & 10 & 1024 & 0.57 & 0.43 & 0.97 & 0.66 \\
16 & 10 & 10 & 2048 & 0.63 & 0.33 & 0.98 & 0.65 \\
16 & 20 & 1 & 128 & 0.33 & 0.13 & 0.80 & 0.42 \\
16 & 20 & 1 & 512 & 0.37 & 0.17 & 0.97 & 0.50 \\
16 & 20 & 1 & 1024 & 0.40 & 0.33 & 1.00 & 0.58 \\
16 & 20 & 5 & 128 & 0.33 & 0.27 & 1.00 & 0.53 \\
16 & 20 & 5 & 512 & 0.47 & 0.27 & 1.00 & 0.58 \\
16 & 20 & 5 & 1024 & 0.70 & 0.43 & 1.00 & 0.71 \\
16 & 20 & 10 & 128 & 0.17 & 0.13 & 1.00 & 0.43 \\
16 & 20 & 10 & 512 & 0.23 & 0.10 & 1.00 & 0.44 \\
16 & 20 & 10 & 1024 & 0.43 & 0.23 & 1.00 & 0.56 \\
18 & 1 & 1 & 128 & 0.43 & 0.33 & 1.00 & 0.59 \\
18 & 1 & 1 & 512 & 0.50 & 0.33 & 1.00 & 0.61 \\
18 & 1 & 1 & 1024 & 0.57 & 0.37 & 1.00 & 0.64 \\
18 & 1 & 1 & 2048 & 0.77 & 0.53 & 1.00 & 0.77 \\
18 & 10 & 1 & 128 & 0.23 & 0.13 & 0.93 & 0.43 \\
18 & 10 & 1 & 512 & 0.53 & 0.33 & 0.98 & 0.62 \\
18 & 10 & 1 & 1024 & 0.20 & 0.20 & 0.77 & 0.39 \\
18 & 10 & 1 & 2048 & 0.33 & 0.30 & 0.95 & 0.53 \\
18 & 10 & 5 & 16 & 0.30 & 0.13 & 0.98 & 0.47 \\
18 & 10 & 5 & 128 & 0.30 & 0.23 & 1.00 & 0.51 \\
18 & 10 & 5 & 512 & 0.43 & 0.23 & 1.00 & 0.56 \\
18 & 10 & 5 & 1024 & 0.53 & 0.40 & 1.00 & 0.64 \\
18 & 10 & 5 & 2048 & 0.73 & 0.53 & 1.00 & 0.76 \\
18 & 10 & 10 & 16 & 0.17 & 0.13 & 1.00 & 0.43 \\
18 & 10 & 10 & 128 & 0.47 & 0.30 & 1.00 & 0.59 \\
18 & 10 & 10 & 512 & 0.63 & 0.37 & 1.00 & 0.67 \\
18 & 10 & 10 & 1024 & 0.67 & 0.40 & 1.00 & 0.69 \\
18 & 10 & 10 & 2048 & 0.70 & 0.53 & 1.00 & 0.74 \\
18 & 20 & 1 & 128 & 0.30 & 0.20 & 0.92 & 0.47 \\
18 & 20 & 1 & 512 & 0.40 & 0.27 & 0.90 & 0.52 \\
18 & 20 & 1 & 1024 & 0.47 & 0.43 & 0.98 & 0.63 \\
18 & 20 & 5 & 16 & 0.27 & 0.17 & 0.75 & 0.39 \\
18 & 20 & 5 & 128 & 0.47 & 0.27 & 0.98 & 0.57 \\
18 & 20 & 5 & 512 & 0.53 & 0.33 & 0.98 & 0.62 \\
18 & 20 & 5 & 1024 & 0.67 & 0.50 & 1.00 & 0.72 \\
18 & 20 & 10 & 16 & 0.27 & 0.17 & 0.97 & 0.47 \\
18 & 20 & 10 & 128 & 0.30 & 0.27 & 1.00 & 0.52 \\
18 & 20 & 10 & 512 & 0.50 & 0.33 & 1.00 & 0.61 \\
18 & 20 & 10 & 1024 & 0.57 & 0.30 & 1.00 & 0.62 \\
20 & 1 & 1 & 128 & 0.50 & 0.27 & 0.98 & 0.58 \\
20 & 1 & 1 & 512 & 0.60 & 0.33 & 1.00 & 0.64 \\
20 & 1 & 1 & 1024 & 0.63 & 0.33 & 1.00 & 0.66 \\
20 & 1 & 1 & 2048 & 0.70 & 0.40 & 1.00 & 0.70 \\
20 & 10 & 1 & 128 & 0.47 & 0.37 & 0.93 & 0.59 \\
20 & 10 & 1 & 512 & 0.40 & 0.43 & 0.97 & 0.60 \\
20 & 10 & 1 & 1024 & 0.33 & 0.20 & 0.88 & 0.47 \\
20 & 10 & 1 & 2048 & 0.37 & 0.37 & 0.85 & 0.53 \\
20 & 10 & 5 & 16 & 0.37 & 0.20 & 1.00 & 0.52 \\
20 & 10 & 5 & 128 & 0.50 & 0.20 & 1.00 & 0.57 \\
20 & 10 & 5 & 512 & 0.40 & 0.23 & 1.00 & 0.54 \\
20 & 10 & 5 & 1024 & 0.57 & 0.23 & 1.00 & 0.60 \\
20 & 10 & 5 & 2048 & 0.67 & 0.30 & 1.00 & 0.66 \\
20 & 10 & 10 & 16 & 0.30 & 0.20 & 1.00 & 0.50 \\
20 & 10 & 10 & 128 & 0.47 & 0.20 & 1.00 & 0.56 \\
20 & 10 & 10 & 512 & 0.37 & 0.23 & 1.00 & 0.53 \\
20 & 10 & 10 & 1024 & 0.60 & 0.30 & 1.00 & 0.63 \\
20 & 10 & 10 & 2048 & 0.60 & 0.40 & 1.00 & 0.67 \\
20 & 20 & 1 & 128 & 0.20 & 0.20 & 0.68 & 0.36 \\
20 & 20 & 1 & 512 & 0.33 & 0.30 & 1.00 & 0.54 \\
20 & 20 & 1 & 1024 & 0.57 & 0.33 & 0.92 & 0.61 \\
20 & 20 & 5 & 16 & 0.43 & 0.30 & 0.90 & 0.54 \\
20 & 20 & 5 & 128 & 0.47 & 0.20 & 1.00 & 0.56 \\
20 & 20 & 5 & 512 & 0.47 & 0.30 & 1.00 & 0.59 \\
20 & 20 & 5 & 1024 & 0.60 & 0.43 & 0.97 & 0.67 \\
20 & 20 & 10 & 16 & 0.33 & 0.23 & 0.97 & 0.51 \\
20 & 20 & 10 & 128 & 0.47 & 0.27 & 1.00 & 0.58 \\
20 & 20 & 10 & 512 & 0.53 & 0.23 & 1.00 & 0.59 \\
20 & 20 & 10 & 1024 & 0.50 & 0.30 & 1.00 & 0.60 \\
22 & 1 & 1 & 128 & 0.40 & 0.17 & 0.98 & 0.52 \\
22 & 1 & 1 & 512 & 0.40 & 0.23 & 1.00 & 0.54 \\
22 & 1 & 1 & 1024 & 0.43 & 0.30 & 1.00 & 0.58 \\
22 & 1 & 1 & 2048 & 0.50 & 0.30 & 1.00 & 0.60 \\
22 & 10 & 1 & 128 & 0.23 & 0.17 & 0.97 & 0.46 \\
22 & 10 & 1 & 512 & 0.23 & 0.33 & 0.95 & 0.51 \\
22 & 10 & 1 & 2048 & 0.40 & 0.17 & 0.90 & 0.49 \\
22 & 10 & 5 & 16 & 0.23 & 0.20 & 1.00 & 0.48 \\
22 & 10 & 5 & 128 & 0.33 & 0.23 & 1.00 & 0.52 \\
22 & 10 & 5 & 512 & 0.47 & 0.33 & 1.00 & 0.60 \\
22 & 10 & 5 & 1024 & 0.47 & 0.20 & 1.00 & 0.56 \\
22 & 10 & 5 & 2048 & 0.50 & 0.27 & 1.00 & 0.59 \\
22 & 10 & 10 & 16 & 0.20 & 0.13 & 1.00 & 0.44 \\
22 & 10 & 10 & 128 & 0.40 & 0.17 & 1.00 & 0.52 \\
22 & 10 & 10 & 512 & 0.40 & 0.20 & 1.00 & 0.53 \\
22 & 10 & 10 & 1024 & 0.40 & 0.23 & 1.00 & 0.54 \\
22 & 10 & 10 & 2048 & 0.43 & 0.30 & 1.00 & 0.58 \\
22 & 20 & 1 & 128 & 0.23 & 0.17 & 0.95 & 0.45 \\
22 & 20 & 1 & 512 & 0.17 & 0.17 & 1.00 & 0.44 \\
22 & 20 & 1 & 1024 & 0.40 & 0.30 & 0.87 & 0.52 \\
22 & 20 & 5 & 16 & 0.30 & 0.13 & 1.00 & 0.48 \\
22 & 20 & 5 & 128 & 0.37 & 0.20 & 1.00 & 0.52 \\
22 & 20 & 5 & 512 & 0.47 & 0.20 & 0.97 & 0.54 \\
22 & 20 & 5 & 1024 & 0.47 & 0.23 & 1.00 & 0.57 \\
22 & 20 & 10 & 16 & 0.33 & 0.13 & 1.00 & 0.49 \\
22 & 20 & 10 & 128 & 0.43 & 0.23 & 1.00 & 0.56 \\
22 & 20 & 10 & 512 & 0.43 & 0.30 & 1.00 & 0.58 \\
22 & 20 & 10 & 1024 & 0.47 & 0.23 & 1.00 & 0.57 \\
24 & 1 & 1 & 16 & 0.20 & 0.17 & 1.00 & 0.46 \\
24 & 1 & 1 & 128 & 0.40 & 0.30 & 1.00 & 0.57 \\
24 & 1 & 1 & 512 & 0.43 & 0.30 & 1.00 & 0.58 \\
24 & 1 & 1 & 1024 & 0.50 & 0.33 & 1.00 & 0.61 \\
24 & 1 & 1 & 2048 & 0.57 & 0.47 & 0.98 & 0.67 \\
24 & 10 & 1 & 512 & 0.37 & 0.20 & 0.97 & 0.51 \\
24 & 10 & 1 & 1024 & 0.20 & 0.10 & 0.93 & 0.41 \\
24 & 10 & 1 & 2048 & 0.27 & 0.27 & 0.93 & 0.49 \\
24 & 10 & 5 & 16 & 0.33 & 0.17 & 1.00 & 0.50 \\
24 & 10 & 5 & 128 & 0.50 & 0.40 & 1.00 & 0.63 \\
24 & 10 & 5 & 512 & 0.60 & 0.47 & 1.00 & 0.69 \\
24 & 10 & 5 & 1024 & 0.57 & 0.47 & 1.00 & 0.68 \\
24 & 10 & 5 & 2048 & 0.53 & 0.37 & 1.00 & 0.63 \\
24 & 10 & 10 & 16 & 0.43 & 0.23 & 1.00 & 0.56 \\
24 & 10 & 10 & 128 & 0.50 & 0.43 & 1.00 & 0.64 \\
24 & 10 & 10 & 512 & 0.37 & 0.40 & 1.00 & 0.59 \\
24 & 10 & 10 & 1024 & 0.57 & 0.50 & 1.00 & 0.69 \\
24 & 10 & 10 & 2048 & 0.60 & 0.47 & 1.00 & 0.69 \\
24 & 20 & 1 & 128 & 0.20 & 0.20 & 0.93 & 0.44 \\
24 & 20 & 1 & 512 & 0.33 & 0.33 & 1.00 & 0.56 \\
24 & 20 & 5 & 16 & 0.30 & 0.23 & 1.00 & 0.51 \\
24 & 20 & 5 & 128 & 0.53 & 0.43 & 1.00 & 0.66 \\
24 & 20 & 5 & 512 & 0.53 & 0.33 & 0.98 & 0.62 \\
24 & 20 & 5 & 1024 & 0.63 & 0.50 & 1.00 & 0.71 \\
24 & 20 & 10 & 16 & 0.37 & 0.20 & 0.98 & 0.52 \\
24 & 20 & 10 & 128 & 0.57 & 0.40 & 1.00 & 0.66 \\
24 & 20 & 10 & 512 & 0.53 & 0.47 & 1.00 & 0.67 \\
24 & 20 & 10 & 1024 & 0.47 & 0.40 & 1.00 & 0.62 \\
26 & 1 & 1 & 16 & 0.13 & 0.13 & 1.00 & 0.42 \\
26 & 1 & 1 & 128 & 0.57 & 0.37 & 1.00 & 0.64 \\
26 & 1 & 1 & 512 & 0.53 & 0.50 & 1.00 & 0.68 \\
26 & 1 & 1 & 1024 & 0.53 & 0.47 & 1.00 & 0.67 \\
26 & 1 & 1 & 2048 & 0.63 & 0.50 & 1.00 & 0.71 \\
26 & 10 & 1 & 512 & 0.27 & 0.23 & 0.95 & 0.48 \\
26 & 10 & 1 & 1024 & 0.13 & 0.10 & 0.93 & 0.39 \\
26 & 10 & 1 & 2048 & 0.37 & 0.33 & 0.95 & 0.55 \\
26 & 10 & 5 & 2 & 0.10 & 0.10 & 0.90 & 0.37 \\
26 & 10 & 5 & 16 & 0.50 & 0.30 & 1.00 & 0.60 \\
26 & 10 & 5 & 128 & 0.57 & 0.37 & 1.00 & 0.64 \\
26 & 10 & 5 & 512 & 0.57 & 0.43 & 1.00 & 0.67 \\
26 & 10 & 5 & 1024 & 0.63 & 0.53 & 1.00 & 0.72 \\
26 & 10 & 5 & 2048 & 0.50 & 0.57 & 1.00 & 0.69 \\
26 & 10 & 10 & 16 & 0.47 & 0.30 & 1.00 & 0.59 \\
26 & 10 & 10 & 128 & 0.53 & 0.37 & 1.00 & 0.63 \\
26 & 10 & 10 & 512 & 0.53 & 0.43 & 1.00 & 0.66 \\
26 & 10 & 10 & 1024 & 0.53 & 0.43 & 1.00 & 0.66 \\
26 & 10 & 10 & 2048 & 0.63 & 0.57 & 0.98 & 0.73 \\
26 & 20 & 1 & 16 & 0.10 & 0.13 & 0.72 & 0.32 \\
26 & 20 & 1 & 512 & 0.23 & 0.20 & 0.90 & 0.44 \\
26 & 20 & 1 & 1024 & 0.40 & 0.33 & 0.95 & 0.56 \\
26 & 20 & 5 & 16 & 0.53 & 0.33 & 1.00 & 0.62 \\
26 & 20 & 5 & 128 & 0.70 & 0.47 & 1.00 & 0.72 \\
26 & 20 & 5 & 512 & 0.60 & 0.47 & 1.00 & 0.69 \\
26 & 20 & 5 & 1024 & 0.53 & 0.30 & 1.00 & 0.61 \\
26 & 20 & 10 & 2 & 0.17 & 0.10 & 0.80 & 0.36 \\
26 & 20 & 10 & 16 & 0.53 & 0.43 & 1.00 & 0.66 \\
26 & 20 & 10 & 128 & 0.57 & 0.30 & 1.00 & 0.62 \\
26 & 20 & 10 & 512 & 0.60 & 0.53 & 1.00 & 0.71 \\
26 & 20 & 10 & 1024 & 0.57 & 0.40 & 1.00 & 0.66 \\
28 & 1 & 1 & 16 & 0.10 & 0.13 & 0.83 & 0.36 \\
28 & 1 & 1 & 128 & 0.47 & 0.33 & 0.93 & 0.58 \\
28 & 1 & 1 & 512 & 0.50 & 0.43 & 0.97 & 0.63 \\
28 & 1 & 1 & 1024 & 0.60 & 0.37 & 0.98 & 0.65 \\
28 & 1 & 1 & 2048 & 0.60 & 0.50 & 1.00 & 0.70 \\
28 & 10 & 1 & 128 & 0.30 & 0.23 & 0.70 & 0.41 \\
28 & 10 & 1 & 1024 & 0.30 & 0.10 & 0.87 & 0.42 \\
28 & 10 & 1 & 2048 & 0.30 & 0.27 & 0.88 & 0.48 \\
28 & 10 & 5 & 16 & 0.37 & 0.40 & 1.00 & 0.59 \\
28 & 10 & 5 & 128 & 0.47 & 0.37 & 0.83 & 0.56 \\
28 & 10 & 5 & 512 & 0.47 & 0.47 & 1.00 & 0.64 \\
28 & 10 & 5 & 1024 & 0.70 & 0.60 & 1.00 & 0.77 \\
28 & 10 & 5 & 2048 & 0.70 & 0.47 & 1.00 & 0.72 \\
28 & 10 & 10 & 16 & 0.30 & 0.30 & 0.98 & 0.53 \\
28 & 10 & 10 & 128 & 0.50 & 0.33 & 0.95 & 0.59 \\
28 & 10 & 10 & 512 & 0.63 & 0.40 & 0.97 & 0.67 \\
28 & 10 & 10 & 1024 & 0.70 & 0.40 & 1.00 & 0.70 \\
28 & 10 & 10 & 2048 & 0.60 & 0.40 & 1.00 & 0.67 \\
28 & 20 & 1 & 512 & 0.27 & 0.23 & 0.95 & 0.48 \\
28 & 20 & 1 & 1024 & 0.27 & 0.10 & 0.93 & 0.43 \\
28 & 20 & 5 & 16 & 0.47 & 0.33 & 0.98 & 0.59 \\
28 & 20 & 5 & 128 & 0.50 & 0.40 & 0.97 & 0.62 \\
28 & 20 & 5 & 512 & 0.50 & 0.37 & 0.92 & 0.59 \\
28 & 20 & 5 & 1024 & 0.50 & 0.37 & 1.00 & 0.62 \\
28 & 20 & 10 & 16 & 0.43 & 0.27 & 0.98 & 0.56 \\
28 & 20 & 10 & 128 & 0.43 & 0.33 & 0.98 & 0.58 \\
28 & 20 & 10 & 512 & 0.60 & 0.40 & 1.00 & 0.67 \\
28 & 20 & 10 & 1024 & 0.50 & 0.30 & 1.00 & 0.60 \\
30 & 1 & 1 & 16 & 0.23 & 0.17 & 1.00 & 0.47 \\
30 & 1 & 1 & 128 & 0.67 & 0.47 & 1.00 & 0.71 \\
30 & 1 & 1 & 512 & 0.73 & 0.57 & 1.00 & 0.77 \\
30 & 1 & 1 & 1024 & 0.70 & 0.63 & 1.00 & 0.78 \\
30 & 1 & 1 & 2048 & 0.73 & 0.73 & 1.00 & 0.82 \\
30 & 10 & 1 & 128 & 0.23 & 0.10 & 0.95 & 0.43 \\
30 & 10 & 1 & 512 & 0.20 & 0.20 & 0.95 & 0.45 \\
30 & 10 & 5 & 16 & 0.53 & 0.40 & 1.00 & 0.64 \\
30 & 10 & 5 & 128 & 0.57 & 0.63 & 1.00 & 0.73 \\
30 & 10 & 5 & 512 & 0.53 & 0.63 & 1.00 & 0.72 \\
30 & 10 & 5 & 1024 & 0.70 & 0.63 & 1.00 & 0.78 \\
30 & 10 & 5 & 2048 & 0.67 & 0.50 & 1.00 & 0.72 \\
30 & 10 & 10 & 16 & 0.43 & 0.27 & 1.00 & 0.57 \\
30 & 10 & 10 & 128 & 0.57 & 0.57 & 1.00 & 0.71 \\
30 & 10 & 10 & 512 & 0.60 & 0.50 & 1.00 & 0.70 \\
30 & 10 & 10 & 1024 & 0.73 & 0.63 & 1.00 & 0.79 \\
\rowcolor{blue!8}
30 & 10 & 10 & 2048 & 0.77 & 0.73 & 1.00 & 0.83 \\
30 & 20 & 1 & 16 & 0.10 & 0.10 & 0.63 & 0.28 \\
30 & 20 & 1 & 128 & 0.17 & 0.13 & 1.00 & 0.43 \\
30 & 20 & 1 & 512 & 0.27 & 0.27 & 1.00 & 0.51 \\
30 & 20 & 1 & 1024 & 0.20 & 0.13 & 0.93 & 0.42 \\
30 & 20 & 5 & 16 & 0.47 & 0.37 & 1.00 & 0.61 \\
30 & 20 & 5 & 128 & 0.53 & 0.57 & 1.00 & 0.70 \\
30 & 20 & 5 & 512 & 0.60 & 0.40 & 1.00 & 0.67 \\
30 & 20 & 5 & 1024 & 0.43 & 0.37 & 1.00 & 0.60 \\
30 & 20 & 10 & 16 & 0.53 & 0.43 & 1.00 & 0.66 \\
30 & 20 & 10 & 128 & 0.53 & 0.43 & 1.00 & 0.66 \\
30 & 20 & 10 & 512 & 0.63 & 0.50 & 1.00 & 0.71 \\
30 & 20 & 10 & 1024 & 0.47 & 0.37 & 1.00 & 0.61
\end{longtable}

\clearpage
\twocolumn